\documentclass[journal]{IEEEtran}
\usepackage{amsmath} % for \boldsymbol
\usepackage{amsthm}
\usepackage{color}
\usepackage{graphicx}
\usepackage{subfigure}
\usepackage{amssymb}
\usepackage{booktabs}
\usepackage{cite}
\usepackage{ulem}
\usepackage{mathrsfs} % Curlicue letters
\usepackage{comment}
\definecolor{darkblue}{rgb}{0.4,0,0.0}
\begin{document}

\title{An Efficient Egocentric Regulator for Continuous Targeting Problems of the Underactuated Quadrotor}

\author{Ziying~Lin,~Wei~Dong,~Sensen~Liu,~Xinjun~Sheng,~and~Xiangyang~Zhu% 
\thanks{The authors are with the State Key Laboratory of Mechanical System
	and Vibration, School of Mechanical Engineering, Shanghai Jiao Tong University, Shanghai 200240, China (e-mail: lynnaero@sjtu.edu.cn; dr.dongwei@sjtu.edu.cn; sensenliu@sjtu.edu.cn; xjsheng@sjtu.edu.cn; mexyzhu@sjtu.edu.cn).}
\thanks{Manuscript received xxx xx, xxxx; revised xxxx xx, xxxx.}}

%\markboth{Journal of Robotics and Automation Letters}%

\maketitle

\begin{abstract}
Flying robots such as the quadrotor could provide an efficient approach for medical treatment or sensor placing of wild animals. In these applications, continuously targeting the moving animal is a crucial requirement. Due to the underactuated characteristics of the quadrotor and the coupled kinematics with the animal, nonlinear optimal tracking approaches, other than smooth feedback control, are required.
However, with severe nonlinearities, it would be time-consuming to evaluate control inputs, and real-time tracking may not be achieved with generic optimizers onboard. To tackle this problem, a novel efficient egocentric regulation approach with high computational efficiency is proposed in this paper. Specifically, it directly formulates the optimal tracking problem in an egocentric manner regarding the quadrotor's body coordinates. Meanwhile, the nonlinearities of the system are peeled off through a mapping of the feedback states as well as control inputs, between the inertial and body coordinates. In this way, the proposed efficient egocentric regulator only requires solving a quadratic performance objective with linear constraints and then generate control inputs analytically. Comparative simulations and mimic biological experiment are carried out to verify the effectiveness and computational efficiency. Results demonstrate that the proposed control approach presents the highest and stablest computational efficiency than generic optimizers on different platforms.
Particularly, on a commonly utilized onboard computer, our method can compute the control action in approximately 0.3 ms, which is on the order of 350 times faster than that of  generic nonlinear optimizers, establishing a control frequency around 3000 Hz. 

\end{abstract}


\IEEEpeerreviewmaketitle

\section{Introduction}
In the past few years, the quadrotor has demonstrated impressive capabilities for biological researches \cite{Cliff2018Robotic,Shah2020penguin}. A recent work also realized the plants' health management via autonomous sensor placement \cite{Farinha2020Placement}. Motivated by those works, the quadrotor equipped with devices such as an anesthesia gun might be able to further apply for wildlife's health management. With autonomous anesthetizing or sensor placing, it may significantly reduce the danger or labor of human beings when coping with agile and aggressive animals \cite{Dodd2010Zoo}.

Different from the case of stationary plants as mentioned in Ref. \cite{Farinha2020Placement}, the moving animal should be continuously targeted before anesthetizing. For such a targeting task, it requires accurate control simultaneously on the position and attitude of the underactuated quadrotor in a continuous-time horizon. Therefore, smooth feedback control \cite{TMECH2020Control,TMECH2020Control2,TMECH2020Control3} is not applicable for such a targeting task, and an effective optimal tracking approach should be formulated. Considering the quadrotor's underactuated characteristics and its coupled kinematics with the moving animal, optimal control of this continuous targeting problem involves severe nonlinearities. Solving such a nonlinear optimal control problem is commonly time consuming. Therefore computational efficiency is a crucial issue regarding the processing capability of onboard computers commonly adopted by quadrotors.

In previous researches, mainly two similar types of optimal problems are extensively studied, i.e., state-to-state maneuver problem \cite{2013computationally,2015Dong,2017movingplatform,Hu2019Time} and optimal tracking multiple fixed waypoints problem \cite{Richter2013Polynomial,Hoffmann2008lines,Cowling2007A,Bouktir2008splines,Sandeepkumar2019FMPC,2018IROS}. For the state-to-state maneuver problem, the computational efficiency is guaranteed since the original problem can be linearizable or the optimal trajectory can be generated offline. For example, in Refs. \cite{2013computationally,2015Dong}, computationally efficient trajectory generation algorithms, as well as subspace stabilization tracking method, are also proposed to maneuver the quadrotor to a time-varying target point in 6-dimensional state space. As the reported problems can be formulated with quadratic or linear optimizations, the optimal trajectories can be generated fairly fast. 
In Refs. \cite{2017movingplatform,Hu2019Time}, time-optimal trajectory generation algorithms are designed for landing a quadrotor onto a moving platform. Although the problem is not formulated as a linear-quadratic problem, it assumes that the platform motion can be represented by a function of time. Therefore, the optimal trajectory can be generated offline without the requirement of computational efficiency.
On the other hand, for the optimal tracking of multiple fixed waypoints problem \cite{Richter2013Polynomial,Hoffmann2008lines,Cowling2007A,Bouktir2008splines,Sandeepkumar2019FMPC,2018IROS}, it is a fully actuated problem that mainly concerns situations in which an invariable terminal target point or a set of fixed waypoints are given. Since the waypoints can be defined a priori, the optimal trajectory can be either generated online or offline, and the computational efficiency is not the primary concern. Instead, the trajectory tracking performance is more crucial \cite{Sandeepkumar2019FMPC,2018IROS}. 

Given that the targeted animal is moving randomly during the whole targeting process, the waypoints cannot be obtained a priori in this underactuated targeting problem, as the fully actuated multiple fixed waypoints problem does. Meanwhile, the real-time tracking problem in this work cannot be linearized directly ,and the animal motion cannot be obtained a priori, as the state-to-state maneuver problem does. This is because three variables, i.e., the pitch angle, the translational position and vertical position of the quadrotor, are controlled simultaneously with two control inputs, i.e., thrust and pitch to ensure the targeting. Meanwhile, the vertical targeting performance objective is determined by the pitch angle, the translational as well as vertical position simultaneously, whereas the pitch angle and the translational position are coupled with each other again. Worse still is that the coupled kinematics with the moving animal further intensifies the nonlinearities of this optimal control problem. The aforesaid coupled nonlinearities and underactuated characteristics exist in the performance objective and the nonlinear constraints, making such a targeting optimal problem unlinearizable. Hence, it is inherently different from the aforementioned linearizable optimization problems such as those investigated in \cite{2013computationally,2015Dong}. Therefore, the optimal tracking methods proposed in these two types of problems are all unapplicable to solve this targeting problem, considering the heavy nonlinearities and computational efficiency simultaneously. 
Meanwhile, according to our outdoor experiments, generic optimizers also cannot directly solve this problem with sufficient computational efficiency. 

To tackle this problem, an efficient egocentric regulator is proposed in this paper. Firstly, regarding the quadrotor's body coordinates, a virtual mathematical model is established in an egocentric manner. In such a way, the decoupling of the pitch angle with the translational position is achieved and the fully actuated system is then obtained. Based on this transformed model, the nonlinearities of the system are peeled off. Then, it only needs to solve a linear quadratic optimization problem and then generate the virtual control inputs analytically. The peeled-off nonlinear part is compensated through a mapping of the feedback states and the virtual inputs between the virtual body and inertial coordinates. This kind of mapping can directly obtain the analytical real control input, achieving high computational efficiency. Mimic biological simulations and experiments are performed to testify the excellent targeting performance and the computational efficiency. Results demonstrate that the proposed control approach presents the highest and stablest computational efficiency than generic optimizers on different platforms. Particularly, it can compute the control action on the order of 350 times faster than generic optimizers on the onboard computer. In such a case, a control frequency of thousands of hertz can be achieved by the proposed approach on a commonly utilized onboard computer.

The main contributions of this work are: 1) To continuously target a moving animal, an efficient egocentric regulator is first proposed, which can present the highest and stablest computational efficiency than generic optimizers on different platforms. Particularly, on the onboard computer,
our method can provide high computational efficiency on the order of 350 times higher than that of generic optimizers; 2) Extensive mimic biological simulations and experiments are first carried out, and the results demonstrate the excellent targeting performance and real-time computational efficiency of the proposed strategy.

The remainder of this paper is organized as follows.  Section II presents the system dynamics and the problem formulation. Section III proposes the efficient egocentric regulator.
Comparative simulations are implemented in Section IV ,while outdoor experiments are conducted in Section V. Section VI concludes the paper.

\section{System Dynamics And Problem Formulation}\label{sec:2}
We first present the general dynamic model of the quadrotor in this section. On this basis, the targeting problem is formulated.

\subsection{Dynamics of quadrotor} 
\begin{figure}
	\centering
	\includegraphics[width=1.0\linewidth]{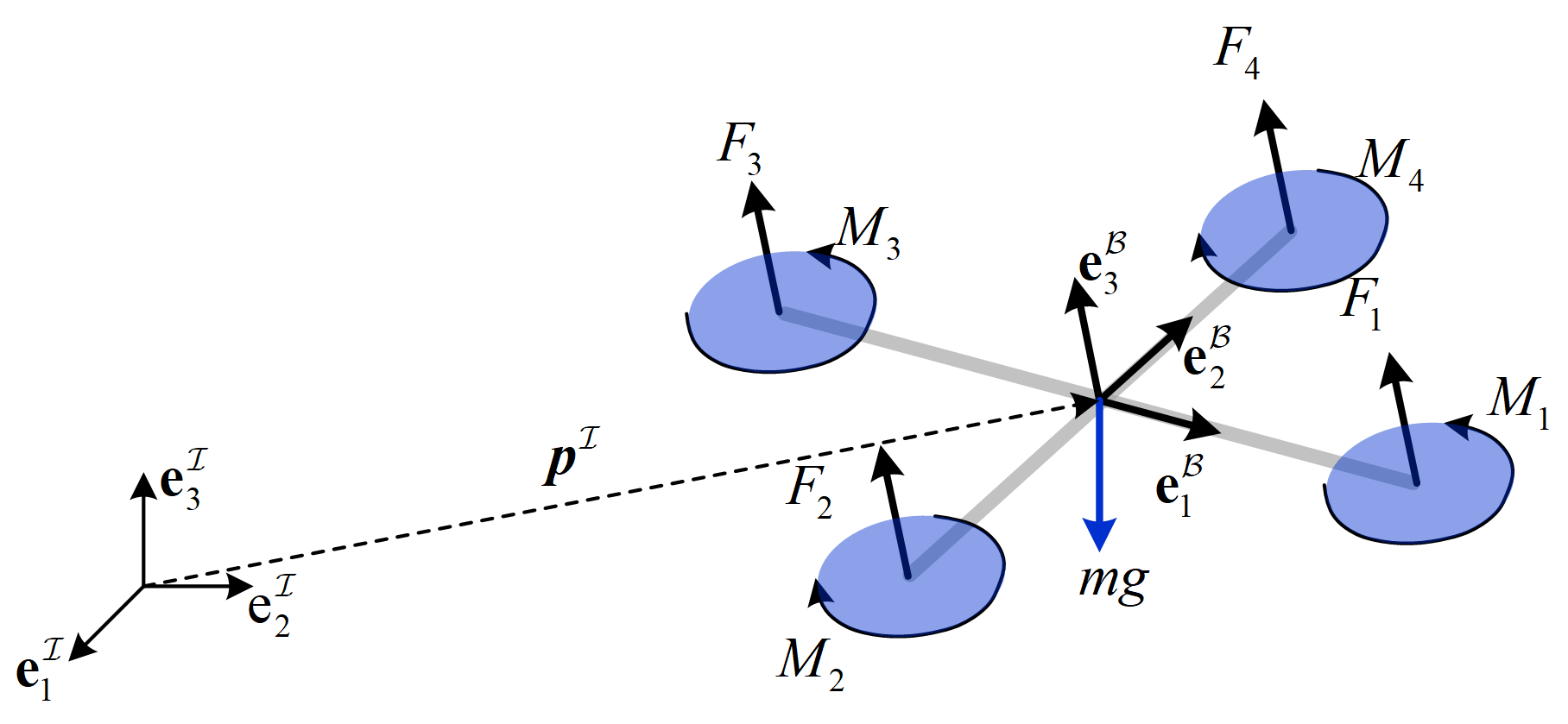}
	\caption{A free body diagram of the quadrotor. $\boldsymbol{e}^{\mathcal{I}}$ is the inertial coordinates and $\boldsymbol{e}^{\mathcal{B}}$ is the body fixed coordinates}
	\label{fig:quad}
\end{figure}

The coordinates and free body diagram of the quadrotor are shown in Fig. \ref{fig:quad}. 

 On this basis, the dynamics of the quadrotor is investigated as follows. First, the force and moment generated by four individual rotors are commonly expressed as
\begin{equation}\label{d2} 
\left\{\begin{array}{ll}
f=\sum F_{i} \quad(i=1,\ldots, 4), &\tau_{x}=\left(F_{4}-F_{2}\right) \cdot L \\
\tau_{z}=M_{1}-M_{2}+M_{3}-M_{4},
&\tau_{y}=\left(F_{3}-F_{1}\right)\cdot L
\end{array}\right.
\end{equation}
where $L$ is the length from the rotor to the center of the mass of the quadrotor, and $F_i$ and $M_i$ are the thrust and torque generated by rotor $i (i \in\{1,2,3,4\})$.

In view of Eq.\eqref{d2}, considering air drag effects, the dynamics of the quadrotor with respect to the inertial coordinates $\boldsymbol{e}^{\mathcal{I}}$ can be expressed as \cite{Dydek2013Adaptive, Dong2014High,Tutorial}
\begin{equation}\label{dyna}
\left\{\begin{array}{ll}
 \dot{\boldsymbol{p}}^\mathcal{I}=\boldsymbol{v}^\mathcal{I} & \dot{\boldsymbol{v}}^\mathcal{I}=\frac{f}{m}_{\mathcal{B}}^\mathcal{I} \boldsymbol{\mathcal{R}}\boldsymbol{e}_{3}^{\mathcal{B}} - g\boldsymbol{e}_{3}^{\mathcal{I}}-\boldsymbol{C}\boldsymbol{v}^\mathcal{I}\\
\dot{\boldsymbol{\Omega}}=\boldsymbol{W} \cdot \boldsymbol{\omega}^\mathcal{B} &  \dot{\boldsymbol{\omega}^\mathcal{B}}=\boldsymbol{J}^{-1} \cdot \boldsymbol{\tau}
\end{array}\right.
\end{equation}
where $\boldsymbol{p}^\mathcal{I}=x_q\boldsymbol{e}_1^{\mathcal{I}}+y_q\boldsymbol{e}_2^{\mathcal{I}}+z_q\boldsymbol{e}_3^{\mathcal{I}}$ denotes the position of the quadrotor in inertial coordinates $\boldsymbol{e}^{\mathcal{I}}$; $\boldsymbol{v}^\mathcal{I} = v_{qx}\boldsymbol{e}_1^{\mathcal{I}}+v_{qy}\boldsymbol{e}_2^{\mathcal{I}}+v_{qz}\boldsymbol{e}_3^{\mathcal{I}}$ is the velocity of the quadrotor in $\boldsymbol{e}^{\mathcal{I}}$;
$m$ is the mass of the quadrotor; 
$g$ is the gravity constant; 
$\boldsymbol{C} = diag\left\{c_{1}, c_2, c_3\right\}$ is the constant matrix to estimate the aerial drag effects in $\boldsymbol{e}^{\mathcal{I}}$;
 $\boldsymbol{\Omega}$ is the attitude, which contains three components: pitch $\theta,$ roll $\phi,$ and yaw $\psi$;
matrix $\boldsymbol{W}$ establishes relationship between attitude velocity $\dot{\boldsymbol{\Omega}}$ and angular velocity $\boldsymbol{\omega}$;
 $\boldsymbol{J}$ is the moment of inertia;
 $\boldsymbol{\tau}=\left[\tau_{x}, \tau_{y}, \tau_{z}\right]^{T}$;$_{\mathcal{B}}^\mathcal{I} \boldsymbol{\mathcal{R}}$ is the transformation matrix from $\boldsymbol{e}^{\mathcal{B}}$ to $\boldsymbol{e}^{\mathcal{I}}$, which is expressed as
 \begin{equation}\label{dynamics}
 _{\mathcal{B}}^\mathcal{I}\boldsymbol{R}=\left[\begin{array}{ccc}
 C_{\psi} C_{\theta} & C_{\psi} S_{\theta} S_{\phi}-S_{\psi} C_{\phi} & C_{\psi} S_{\theta} C_{\phi}+S_{\psi} S_{\phi} \\
 S_{\psi} C_{\theta} & S_{\psi} S_{\theta} S_{\phi}+C_{\psi} C_{\phi} & S_{\psi} S_{\theta} C_{\phi}-C_{\psi} S_{\phi} \\
 -S_{\theta} & C_{\theta} S_{\phi} & C_{\theta} C_{\phi}
 \end{array}\right]
 \end{equation}
 in which $S_{x}=\sin (x) \text { and } C_{x}=\cos (x)$.

\subsection{Problem Formulation}
To clarify the formulation of the optimal control strategy designed in this work, three assumptions are
 imposed as follows.

\noindent {\it{\textbf{{Assumption}}}} \textbf{1}. 
{\it The pitch angle $\theta$ and the roll angle $\phi$ of the quadrotor can be set directly without dynamics and delay such that they can be regarded as the control inputs.}

\noindent {\it{\textbf{{Assumption}}}} \textbf{2}. 
{\it Considering that the yaw angle $\psi$ of the quadrotor can be controlled separately without underactuated characteristics, it can be approximated as zero, i.e., $\psi = 0$.}

\noindent {\it{\textbf{{Assumption}}}} \textbf{3}. 
{\it{The animal's velocity is regarded as a constant in each controller updating timestep.}}

 In view of the developed model, the control objective can be described as follows. An optimal control input of the quadrotor is required to design in such a way the quadrotor can continuously target the animal and keep in a safe distance. The targeting problem is illustrated in Fig. \ref{fig:model2}, where $\boldsymbol{p}_t^\mathcal{I}=x_t\boldsymbol{e}_1^{\mathcal{I}}+y_t\boldsymbol{e}_2^{\mathcal{I}}+z_t\boldsymbol{e}_3^{\mathcal{I}}$ denotes the position of the desired-point in inertial coordinate $\boldsymbol{e}^{\mathcal{I}}$. 
 
%Then, the state vectors of the quadrotor and the animal can be denoted as $\boldsymbol{x}^{q}=\left[x_{q}, y_{q}, z_{q}, {v}_{qx},{v}_{qy}, {v}_{qz}\right]^{T}, \boldsymbol{x}^{t}=\left[x_{t}, y_{t}, z_{t}, {v}_{tx},{v}_{ty}, {v}_{tz}\right]^{T}$, respectively. 
%To facilitate the problem formulation, firstly, define the position and the velocity of the desired-point as $\boldsymbol{p}_t^\mathcal{I}=x_t\boldsymbol{e}_1^{\mathcal{I}}+y_t\boldsymbol{e}_2^{\mathcal{I}}+z_t\boldsymbol{e}_3^{\mathcal{I}}, \boldsymbol{v}_t^\mathcal{I} = v_{tx}\boldsymbol{e}_1^{\mathcal{I}}+v_{ty}\boldsymbol{e}_2^{\mathcal{I}}+v_{tz}\boldsymbol{e}_3^{\mathcal{I}}$ in inertial coordinates $\boldsymbol{e}^{\mathcal{I}}$. 

 Combining with Assumption 1, we can define the state vector $\boldsymbol{x}$ and the control input $\boldsymbol{u}$ of the system as:
\begin{equation}\label{4}
\begin{aligned}
&\boldsymbol{x}=[\left(\boldsymbol{p}^\mathcal{I}-\boldsymbol{p}_t^\mathcal{I}\right)^T,\left(\boldsymbol{v}^\mathcal{I}-\boldsymbol{v}_t^\mathcal{I}\right)^T,\left(\boldsymbol{v}^\mathcal{I}\right)^T]^T,\boldsymbol{u} = [\frac{f}{m},\theta,\phi]^T\\
\end{aligned}
\end{equation}
where $\boldsymbol{v}_t^\mathcal{I} = v_{tx}\boldsymbol{e}_1^{\mathcal{I}}+v_{ty}\boldsymbol{e}_2^{\mathcal{I}}+v_{tz}\boldsymbol{e}_3^{\mathcal{I}}$ denotes
the velocity of the desired-point in inertial coordinate $\boldsymbol{e}^{\mathcal{I}}$. The quadrotor's velocity $\boldsymbol{v}^\mathcal{I}$ is considered as separate states. It is separately introduced to reflect the air drag effect. In this way, the static case and the uniform motion case, where the animal and the quadrotor are both relatively stationary, can be distinguished.

%To facilitate the optimal control, it can be assumed that the air drag is proportional to the quadrotor's translational velocity. 
%Meanwhile, the animal's velocity is regarded as a constant in each controller updating timestep. Then, based on Eq. \eqref{dyna}, the dynamics of the targeting process can then be reformulated as follows:
Based on Assumption 2, Assumption 3 and Eq. \eqref{dyna}, the dynamics of the targeting process can then be reformulated as follows:
\begin{figure}
	\centering
	\includegraphics[width=0.9\linewidth]{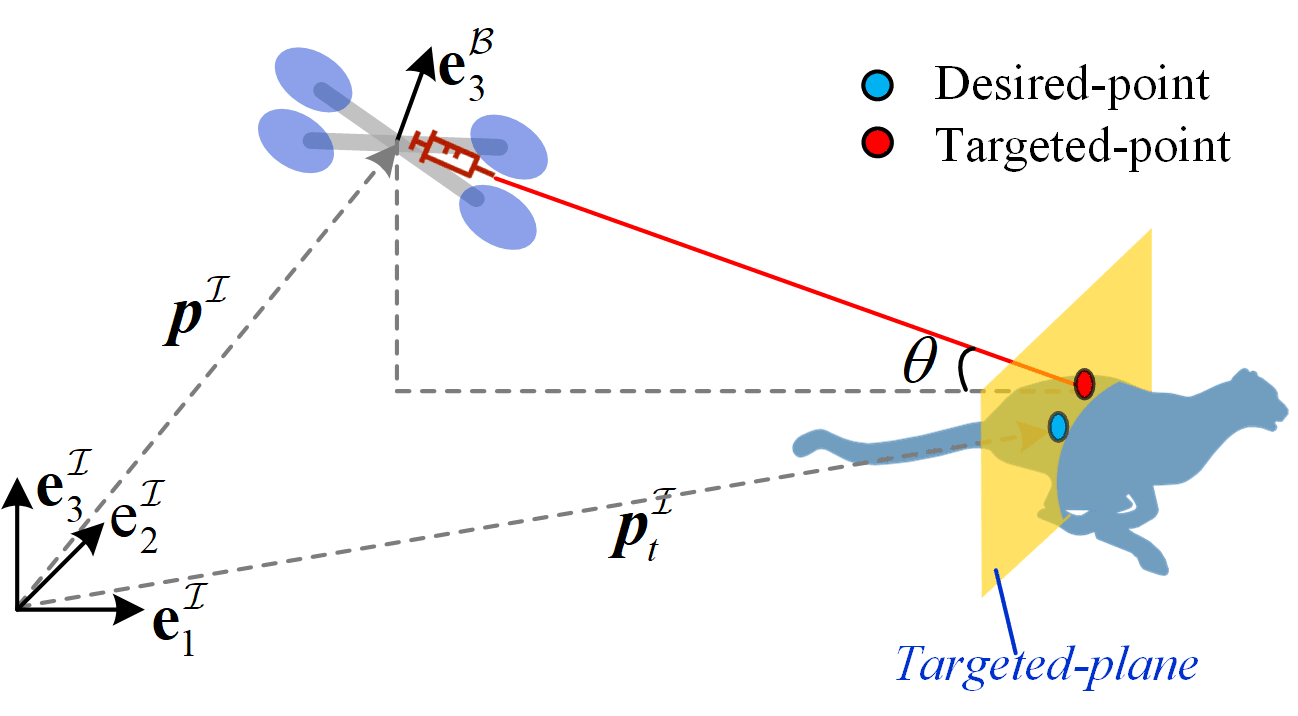}
	\caption{A schematic diagram for the targeting problem. Desired-point is the desired hit point of the anethesia gun in the animal; targeted plane denotes the plane which is parallel to the $\boldsymbol{e}_2^\mathcal{I}\boldsymbol{e}_3^\mathcal{I}$ plane while passing through the desired-point; targeted-point is the intersection of the anesthesia gun's extension line and the targeted plane.}
	\label{fig:model2}
\end{figure}
\begin{equation} \label{1}
\begin{aligned}
&\boldsymbol{\dot{x}}(t)=\boldsymbol{h}\left[\boldsymbol{x}(t),\boldsymbol{u}(t)\right]=[{\boldsymbol{x}_{4-6}}^T, \boldsymbol{a}^T, \boldsymbol{a}^T]^{T}
\end{aligned}
\end{equation}
where $\boldsymbol{x}_{i}$ denotes the $i$-th component of $\boldsymbol{x}$; $\boldsymbol{a} = [ \boldsymbol{u}_1C_{\boldsymbol{u}_3}S_{\boldsymbol{u}_2}-c_1\boldsymbol{x}_7, -\boldsymbol{u}_1S_{\boldsymbol{u}_3}-c_2\boldsymbol{x}_8,\boldsymbol{u}_1C_{\boldsymbol{u}_3}C_{\boldsymbol{u}_2}-c_3\boldsymbol{x}_9-g]^T$, where $ \boldsymbol{u}_{i}$ denotes $i$-th component of $\boldsymbol{u}$.

To facilitate the optimal control, Define $d_x \triangleq x_{t}-x_{q}$ as the distance from the center of mass of the quadrotor to the targeted plane, and define $d_y \triangleq y_{t}-y_{q}$, $d_z \triangleq (d_x\tan\theta-(z_{q}-z_{t}))$ as the distance along $\boldsymbol{e}_2^\mathcal{I}, \boldsymbol{e}_3^\mathcal{I}$ between the targeted-point and the desired-point on the targeted plane. For this targeting task, the control objective is to design an optimal control input $\boldsymbol{u}$ such that $d_x$ is as close to the safe distance (3 meters in this paper) as possible. At the same time, $d_y, d_z$ and control input should be minimized within a finite time horizon $t\in[0,t_f]$.
% For simplicity, it is assumed that the pitch angle $\boldsymbol{u}_2$ is small, satisfying $\tan(\boldsymbol{u}_2) \approx \boldsymbol{u}_2$. Hence, $d_z$ can be reformulated as: $d_z = -\boldsymbol{x}_{1}\boldsymbol{u}_{2}-\boldsymbol{x}_2$.
  In this way, the considered optimal control problem of the targeting problem can be formulated as the follows:
\begin{equation}\label{function}
\begin{aligned}
\min \quad & \displaystyle \mathcal{J}=\int_{0}^{{{t}_{f}}} G[\boldsymbol{x}(t), \boldsymbol{u}(t),t]\text{d}t\\
\textrm{s.t.} \quad   &     \displaystyle\boldsymbol{\dot{x}}(t)=\boldsymbol{h}[\boldsymbol{x}(t),\boldsymbol{u}(t)],\boldsymbol{x}(0)=\boldsymbol{{\xi}_{0}}  \\
\end{aligned}
\end{equation}
where  the terminal time $t_{f}$ is fixed; $\displaystyle G[\boldsymbol{x}(t), \boldsymbol{u}(t),t] ={\frac{1}{2}\boldsymbol{u}^T\boldsymbol{u}+k_{1}{{\left( d_x-3 \right)}^{2}}+k_2d_{y}^2+k_3d_{z}^2}$; $\boldsymbol{{\xi}_{0}} \in \mathbb{R}^{9}$ is the initial value of the state vector, which is known a priori; $k_{i}, i = 1\ldots3$ are the positive weights.

From \eqref{function}, it can be inferred that the pitch angle $\boldsymbol{u}_2$ is coupled with the translational position $\boldsymbol{x}_1$ in the performance objective $\mathcal{J}$. Additionally, there exists severe nonlinear characteristics in this optimal control problem, owing to the coupled kinematics with the animal and the inherent nonlinearities of the quadrotor.
 Hence, this optimal control problem can neither be linearized nor be solved analytically. The general ways of solving a nonlinear optimal control problem is to solve it numerically with generic optimizers. Due to the coupled and nonlinear characteristics, directly solving this nonlinear optimal control problem with generic optimizers is time-consuming and cannot meet the real-time calculation requirements for real system.
\section{Efficient Egocentric Regulator}\label{sec:3}
\begin{figure}
	\centering
	\includegraphics[width=1\linewidth]{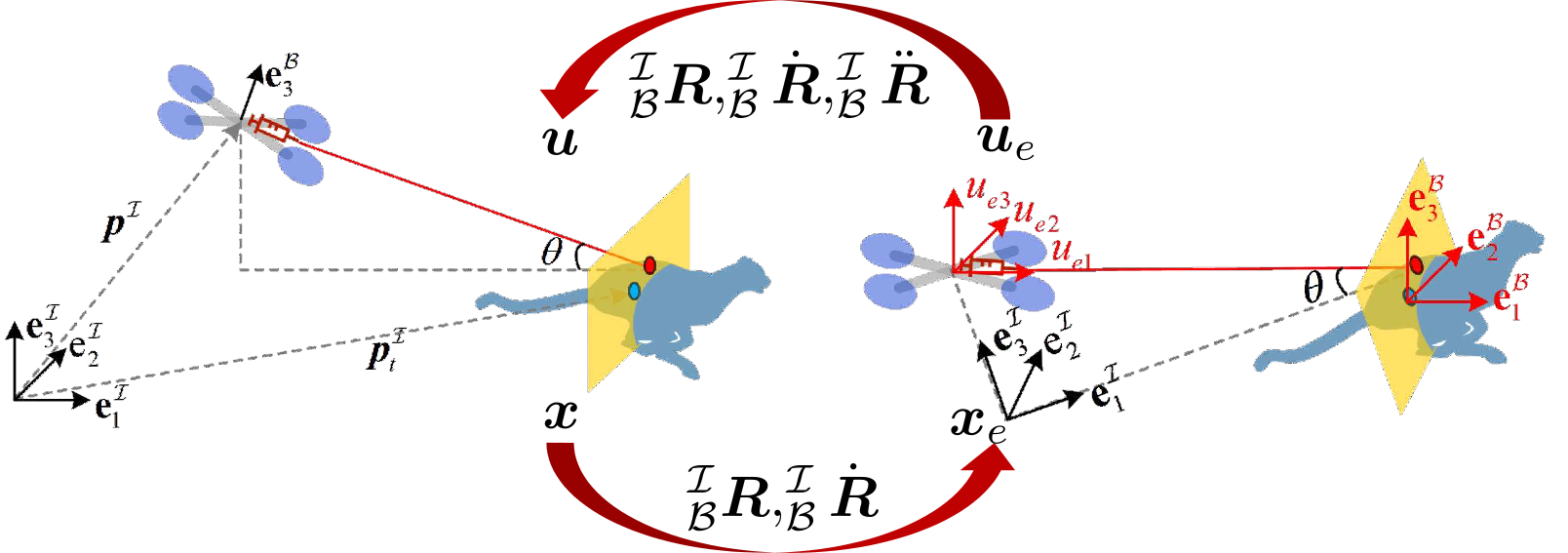}
	\caption{The illustration of the EER. As shown in the right part, a virtual coordinate $\mathcal{V}$ parallel to the quadrotor's instantaneous body coordinate $\boldsymbol{e}^{\mathcal{B}}$ is established while taking the desired-point as the origin. Based on this, a virtual system model is established, where the virtual state $\boldsymbol{x}_e$ is obtained by mapping $\boldsymbol{x}$ from the inertial coordinate $\boldsymbol{e}^{\mathcal{I}}$ to the virtual coordinate. Then, a linear quadratic optimization problem is built up to generate the virtual optimal control input $\boldsymbol{u}_e$ analytically. Finally, the real control input $\boldsymbol{u}$ can be derived by mapping $\boldsymbol{u}_e$ back to $\boldsymbol{e}^{\mathcal{I}}$.}
	\label{fig:virtual}
\end{figure}

To tackle this issue, a novel method called  Efficient Egocentric Regulator (EER) with high computational efficiency is proposed in this section, which peels off the nonlinearities from the optimal control problem \eqref{function}, as shown in Fig. \ref{fig:virtual}.

Firstly, a virtual coordinate $\mathcal{V}$ parallel to the quadrotor's instantaneous body coordinate ${\mathcal{B}}$ is established while taking the desired-point as the origin. Then, the virtual position and the velocity of the quadrotor in virtual coordinate $\boldsymbol{e}^\mathcal{V}$ can be defined as $
 \boldsymbol{p}^\mathcal{V} = x_e\boldsymbol{e}_1^{\mathcal{V}}+y_e\boldsymbol{e}_2^{\mathcal{I}}+z_e\boldsymbol{e}_3^{\mathcal{V}}, \boldsymbol{v}^\mathcal{V} = v_{xe}\boldsymbol{e}_1^{\mathcal{V}}+v_{ye}\boldsymbol{e}_2^{\mathcal{I}}+v_{ze}\boldsymbol{e}_3^{\mathcal{V}}$. The virtual state $\boldsymbol{x}_e$ and the virtual control input $\boldsymbol{u}_e$ of the considered system in virtual coordinate can be defined as 
\begin{equation}\label{vi}
 \boldsymbol{x}_e=\left[x_{e}+r^*, y_{e},  z_{e},v_{xe},v_{ye}, v_{ze}\right]^{T},\boldsymbol{u}_e=\left[u_{e1}, u_{e2}, u_{e3}\right]^{T}
\end{equation}
 where $r^*$ is a constant, denoted as the instantaneously desired position of $x_e$; $u_{e1}$, $u_{e2}$ and $u_{e3}$ are the accelerations in virtual coordinate along $\boldsymbol{e}_1^\mathcal{V}$, $\boldsymbol{e}_2^\mathcal{V}$ $\boldsymbol{e}_3^\mathcal{V}$. 
 
 \noindent {\it{\textbf{{Theoreom}}}} \textbf{1}. 
 {\it{The mapping of the feedback states and control inputs, between the inertial and virtual coordinates can be approximated as}}
 \begin{equation}\label{mapping}
\begin{aligned}
\boldsymbol{x}_e = [\boldsymbol{I}_{2\times2}\otimes(_{\mathcal{B}}^\mathcal{I}\boldsymbol{\mathcal{R}})^{T}] \boldsymbol{x}_{1-6}+[r^*,\boldsymbol{0}_{1\times5}^T]^T,\boldsymbol{u}_e= _{\mathcal{B}}^\mathcal{I}\boldsymbol{\mathcal{R}}^{T}{\boldsymbol{a}}
\end{aligned}
\end{equation}
{\it where $\boldsymbol{x}_{1-6}$ denotes the first six components of $\boldsymbol{x}$.}
\begin{proof}
It can be easily obtained that 
\begin{equation}
\begin{aligned}\label{transform1}
&\boldsymbol{p}^\mathcal{V} =_{\mathcal{B}}^\mathcal{I}\boldsymbol{\mathcal{R}}^{T} \boldsymbol{x}_{1-3}
\end{aligned}
\end{equation}
where $\boldsymbol{x}_{1-3}$ denotes the first three components of $\boldsymbol{x}$. 
Taking the first and second time derivative of Eq. \eqref{transform1} gives
\begin{equation}\label{transform2}
\begin{aligned}
&\boldsymbol{v}^\mathcal{V} =_\mathcal{B}^\mathcal{I}\dot{\boldsymbol{\mathcal{R}}}\boldsymbol{x}_{1-3}+_\mathcal{B}^\mathcal{I}\boldsymbol{\mathcal{R}}^{T} \dot{\boldsymbol{x}}_{1-3} \\
&\boldsymbol{u}_e=_\mathcal{B}^\mathcal{I}\ddot{\boldsymbol{\mathcal{R}}}^{T} \boldsymbol{x}_{1-3}+2 _\mathcal{B}^\mathcal{I}\dot{\boldsymbol{\mathcal{R}}}^{T} \dot{\boldsymbol{x}}_{1-3}+_\mathcal{B}^\mathcal{I}\boldsymbol{\mathcal{R}}^{T} \ddot{\boldsymbol{x}}_{1-3}
\end{aligned}
\end{equation}

Furthermore, it can be assumed that $_\mathcal{B}^\mathcal{I}\dot{\boldsymbol{\mathcal{R}}}^{T}\approx\boldsymbol{0}_{3\times3},_\mathcal{B}^\mathcal{I}\ddot{\boldsymbol{\mathcal{R}}}^{T}\approx\boldsymbol{0}_{3\times3}$ in each controller updating timestep. Substituting this assumption into Eq. \eqref{transform1} and Eq. \eqref{transform2}, one can then prove Theoreom 1.
\end{proof}

 \noindent {\it{\textbf{{Theoreom}}}} \textbf{2}. {\it The optimal control input $\boldsymbol{u}^*$ given by the proposed EER in inertial coordinate is}
\begin{equation}
\begin{aligned}
\boldsymbol{u}^{*} & =\left[\left\|{\boldsymbol{a}}^*+g\boldsymbol{e}_{3}^{\mathcal{I}}-\boldsymbol{C}\boldsymbol{x}_{4-6}\right\|,\arctan\left(\frac{{\boldsymbol{a}}_{1}^*+c_1{\boldsymbol{x}}_{4}}{{\boldsymbol{a}}_{3}^*+c_3{\boldsymbol{x}}_{6}+g}\right)\right.,\\ &\left.\arcsin\left(\frac{-{\boldsymbol{a}}_{2}^*-c_2{\boldsymbol{x}}_{5}}{\boldsymbol{u}_{1}^*}\right)\right]^T \\
\end{aligned}
\end{equation}
{\it {where $\boldsymbol{a}^*=_{\mathcal{B}}^\mathcal{I}\boldsymbol{\mathcal{R}}\boldsymbol{K}\boldsymbol{x}_e$ is the optimal acceleration in inertial coordinate $\mathcal{I}$, in which $\boldsymbol{K}\in\mathbb{R}^{3\times6}$ is a control gain matrix.}}
 
\begin{proof}
 Firstly, based on the definition in Eq. \eqref{vi}, the virtual system model can be established in a egocentric manner as:
\begin{equation} \label{FER}
\dot{\boldsymbol{x}}_{e}=\boldsymbol{ A}_e\boldsymbol{x}_e+\boldsymbol{B}_e \boldsymbol{u}_e 
\end{equation}
with $\boldsymbol{A}_e=\begin{bmatrix}
\boldsymbol{0}_{3\times3}&\boldsymbol{I}_{3\times3}\\
\boldsymbol{0}_{3\times3}&\boldsymbol{0}_{3\times3}
\end{bmatrix}$, $\boldsymbol{B}_e = \left[\boldsymbol{0}_{3\times3},\boldsymbol{I}_{3\times3}\right]^T$.
Based on this transformed model, the pitch angle is decoupled with the translational position and a fully actuated linear system with three inputs and three outputs is obtained. In addition, the nonlinearities of the system are peeled off and the separated nonlinear part can be compensated through a mapping of the feedback states and control inputs, between the inertial and virtual coordinates.

Then, it only needs to solve a linear quadratic optimization problem.
For the targeting task, the goal is to design an optimal controller which minimizes the sum of the targeting error and input energy. Based on the virtual coordinate, a quadratic optimal control problem can be formulated as 
\begin{equation} 
\begin{aligned}\label{LQR}
\min\quad&\mathcal{J}_{e}=\frac{1}{2} \int_{0}^{\infty} \boldsymbol{x}_e^{T} \boldsymbol{Q}_{1} \boldsymbol{x}_e+\boldsymbol{u}_e^{T} \boldsymbol{Q}_{2} \boldsymbol{u}_e \mathrm{d}t\\
s.t.\quad& \dot{\boldsymbol{x}}_e=\boldsymbol{ A}_e\boldsymbol{x}_e+\boldsymbol{B}_e \boldsymbol{u}_e, \boldsymbol{x}_e(0) = \boldsymbol{x}_{e0}
\end{aligned}
\end{equation}
where $\boldsymbol{Q}_1\in\mathbb{R}^{6\times6}$ denotes semi-definite diagonal state weighting matrix; $\boldsymbol{Q}_2 \in \mathbb{R}^{3\times3}$ is a positive definite diagonal control weighting matrix; $\boldsymbol{x}_{e0}\in \mathbb{R}^{9}$ is the initial state of $\boldsymbol{x}_e$. Then, by solving such a linear quadratic optimization problem, the optimal virtual control input $\boldsymbol{u}_e^{*}$ can be obtained:
\begin{equation}\label{eq:12}
\begin{aligned}
&\boldsymbol{u}_e^{*}=\boldsymbol{K}\boldsymbol{x}_e(t)\\
% &\boldsymbol{x}_e^{*}=\boldsymbol{x}_{e0}\exp\left\{\left(\boldsymbol{A}_e-\boldsymbol{B}_e\boldsymbol{Q}_{2}^{-1} \boldsymbol{B}_e^{T} \boldsymbol{P}\right)t\right\}\\
\end{aligned}
\end{equation}
where $\boldsymbol{x}_e$ can be obtained from the mapping given in  Eq.\eqref{mapping}; $\boldsymbol{K} = -\boldsymbol{Q}_{2}^{-1} \boldsymbol{B}_e^{T} \boldsymbol{P}$, $\boldsymbol{P}\in\mathbb{R}^{6\times6}$ is a positive constant matrix found by solving the continuous time algebraic Riccati equation $-\boldsymbol{P} \boldsymbol{A}_e-\boldsymbol{A}_e^{T} \boldsymbol{P}+\boldsymbol{PB}_e \boldsymbol{Q}_{2}^{-1} \boldsymbol{B}_e^{T} \boldsymbol{P}-\boldsymbol{Q}_{1}=0$. As $\boldsymbol{Q}_1$ and $\boldsymbol{Q}_2$ are constant, $\boldsymbol{P}$ can be solved offline. In this work, to be consistent with the cost function given in Eq. \eqref{function}, $r^*$ in $\boldsymbol{x}_e$ is selected with $r^*=3/\cos\theta$, and it can be assumed as a constant in each controller updating timestep considering high computational efficiency. In practice, it is also feasible to directly select $r^*=3$.

% \begin{equation}
% \begin{aligned}
% &\boldsymbol{x}_{4-6}^* = _{\mathcal{B}}^\mathcal{I}\boldsymbol{\mathcal{R}}\boldsymbol{x}_{e,4-6}^{*},{\boldsymbol{a}}^* = _{\mathcal{B}}^\mathcal{I}\boldsymbol{\mathcal{R}}\boldsymbol{u}_e^{*}\\
% \end{aligned}
% \end{equation}

After calculating the optimal virtual control input, combined with Theoreom 1, the desired acceleration in inertial coordinate $\mathcal{I}$ can be obtained as
\begin{equation}\label{eq:13}
\begin{aligned}
&{\boldsymbol{a}}^* = _{\mathcal{B}}^\mathcal{I}\boldsymbol{\mathcal{R}}\boldsymbol{u}_e^{*} = _{\mathcal{B}}^\mathcal{I}\boldsymbol{\mathcal{R}}\boldsymbol{K}\boldsymbol{x}_e\\
\end{aligned}
\end{equation}

Therefore, according to the system dynamics in Eq. \eqref{1}, the optimal control input $\boldsymbol{u}^*$ in inertial coordinate can be derived analytically as Theoreom 2.
\end{proof}

\noindent {\it{\textbf{{Corollary}}}} \textbf{1.} {\it When $\sin\phi\approx0$, the control of the EER in $\boldsymbol{e}_2^{\mathcal{I}}$ can be equivalent to a PD controller.}
\begin{proof}
According to the definition of $\boldsymbol{Q}_{2}, \boldsymbol{B}_e$, and $\boldsymbol{P}$, the control gain matrix $\boldsymbol{K}$ can be rewritten as
\begin{equation}
\begin{aligned}
&\boldsymbol{K} = -\boldsymbol{Q}_{2}^{-1} \boldsymbol{B}_e^{T} \boldsymbol{P} = \left[\boldsymbol{K}_1,\boldsymbol{K}_2\right]\\
\end{aligned}
\end{equation}
where $\boldsymbol{K}_1\in\mathbb{R}^{3\times3},\boldsymbol{K}_2\in\mathbb{R}^{3\times3}$ are two diagnoal matrixes.

Based on Eq. \eqref{mapping} and Eq. \eqref{eq:13}, one can obtain
\begin{equation}
\begin{aligned}
{\boldsymbol{a}}^* &= _{\mathcal{B}}^\mathcal{I}\boldsymbol{\mathcal{R}}\left[\boldsymbol{K}_1,\boldsymbol{K}_2\right]\left(\boldsymbol{I}_{2\times2}\otimes(_{\mathcal{B}}^\mathcal{I}\boldsymbol{\mathcal{R}})^{T} \boldsymbol{x}_{1-6}+[r^*,\boldsymbol{0}_{1\times5}^T]^T\right)\\
& = \left[_{\mathcal{B}}^\mathcal{I}\boldsymbol{\mathcal{R}}{\boldsymbol{K}_1}_{\mathcal{B}}^\mathcal{I}\boldsymbol{\mathcal{R}}^T, _{\mathcal{B}}^\mathcal{I}\boldsymbol{\mathcal{R}}{\boldsymbol{K}_2}_{\mathcal{B}}^\mathcal{I}\boldsymbol{\mathcal{R}}^T\right]\boldsymbol{x}_{1-6} + _{\mathcal{B}}^\mathcal{I}\boldsymbol{\mathcal{R}}[\boldsymbol{K}_{1,1}r^*, 0, 0]^T\\
\end{aligned}
\end{equation}
where $\boldsymbol{K}_{1,1}$ denotes the first diagonal element of $\boldsymbol{K}_1$. Given the near hovering state, i.e., $\sin\phi\approx0$ and according to Eq. \eqref{dynamics}, the optimal accelaration $\boldsymbol{a}_2^*$ along $\boldsymbol{e}_2^{\mathcal{I}}$ can be given by
\begin{equation}
\begin{aligned}
{\boldsymbol{a}}_2^* & = 
\begin{bmatrix}
0&\boldsymbol{K}_{1,2}C_{\phi}^2&0&0&\boldsymbol{K}_{2,2}C_{\phi}^2&0
\end{bmatrix}\boldsymbol{x}_{1-6}\\
& \triangleq k_p\boldsymbol{x}_2 + k_d\boldsymbol{x}_4\\
\end{aligned}
\end{equation}
where  $\boldsymbol{K}_{1,2}, \boldsymbol{K}_{2,2}$ denote the second diagonal element of $\boldsymbol{K}_1$ and $\boldsymbol{K}_2$, respectively; $k_p, k_d$ are the control gains. Therefore, it can be inferred that the control of the EER in $\boldsymbol{e}_2^{\mathcal{I}}$ can be equivalent to a PD controller.
\end{proof}

\noindent{\it{\textbf{{Remark}}} \textbf{1.} {\it From corollary 1, it can be inferred that the position control along $\boldsymbol{e}_2^{\mathcal{I}}$ can be independent of the proposed EER control and then simplified as an independent PD control. In this case, the state variable $\boldsymbol{x_e}$ 
can be reduced to 4 dimensions, and the dimensions of other parameters will also be reduced accordingly.}}
\section{Numerical Evaluation}
In this section, to verify the effectiveness of this work, numerical simulations are first carried out.
%In this section, to effectively verify the computational efficiency and excellent targeting performance of the proposed EER, both EER and generic optimizers (BVP and GPM) are implemented in simulation tests and onboard tests.
\subsection{Generic optimizers}
Generally, to directly tackle the optimal control problem in Eq. \eqref{function}, there are two types of approaches, i.e., the indirect one and the direct one \cite{Benson2006GPM}. In this paper, for comparison, two representative methods are selected from each type, that is, the Gauss Pseudospectral Method\cite{Benson2006GPM, Rao2010GPOPS} and the Boundary Value Problem method \cite{Survey}. 
\subsubsection{Gauss pseudospectral method}
Firstly, Legendre-Gauss (LG) points were selected as interpolation nodes and the values of the state and the control variables on LG points are regarded as unknown parameters. On this basis, the state and control trajectories are approximated using Lagrange interpolating polynomials and the system dynamics can be obtained in a discretized manner. Furthermore, the continuous cost function of Eq. \eqref{function} is approximated using a Gauss quadrature. Then, the optimal control problem can be transformed into a nonlinear programming problem (NLP). Due to the space limitation, the specific discretization process of the optimal trajectory control problem via GPM is omitted here. Detailed principles of GPM can be found in \cite{Benson2006GPM,Rao2010GPOPS}. The resulting transcribed NLP via GPM can be written as follows:
\begin{equation}
\begin{aligned}\label{Eq:16}
	\min\quad\displaystyle &\mathcal{J}=\frac{t_{f}}{2} \boldsymbol\omega^{T}\left[\frac{1}{2}\boldsymbol{\eta_{1}}^{2}+\frac{1}{2} \boldsymbol{\eta_{2}}^{2}+k_{1}\left(-\boldsymbol{\xi_{1}}-3 \cdot\boldsymbol{1}_{N\times1}\right)^{2} \right. \\
	&\quad\left.+k_2\boldsymbol{\xi_{2}}^2 + k_3\left((-\boldsymbol{\xi_{1}}\tan\boldsymbol{\eta_{3}}-\boldsymbol{\xi_{3}})^2\right)\right] \\
s.t.\quad&\boldsymbol{D}\left[\boldsymbol{\xi_{0,i}},\boldsymbol{\xi_{i}}^{T}\right]^{T}-\frac{t_{f}}{2} \boldsymbol{\xi_{i+3}}=0,  i=1\ldots3 \\
&\boldsymbol{D}\left[\boldsymbol{\xi_{0,i}}, \boldsymbol{\xi_{i}}^{T}\right]^{T}-\frac{t_{f}}{2}\left(\boldsymbol{\eta_1} C_{\boldsymbol{\eta_3}} S_{\boldsymbol{\eta_2}}-c_{1} \boldsymbol{\xi_{7}}\right)=0, i=4,7 \\
&\boldsymbol{D}\left[\boldsymbol{\xi_{0,i}}, \boldsymbol{\xi_{i}}^{T}\right]^{T}-\frac{t_{f}}{2}\left(-\boldsymbol{\eta_1} S_{\boldsymbol{\eta_3}}-c_{2} \boldsymbol{\xi_{8}}\right)=0,i=5,8 \\
&\boldsymbol{D}\left[\boldsymbol{ \xi_{0,i}}, \boldsymbol{\xi_{i}}^{T}\right]^{T}-\frac{t_{f}}{2}\left(\boldsymbol{\eta_{1}} C_{\boldsymbol{\eta_3}} C_{\boldsymbol{\eta_2}}-g \boldsymbol{1}-c_{3} \boldsymbol{\xi_{9}}\right)=0,\\
&i=6,9
\end{aligned}
\end{equation}
where
$\boldsymbol{\omega}$ is the Gauss weight vector;
$\displaystyle\boldsymbol{\eta}_j\in\mathbb{R}^{N},j=1\ldots3$ is the discretized vector, composed of all values of $\boldsymbol{u}_j$ at LG points;
Similarly, $\displaystyle\boldsymbol{\xi}_i\in\mathbb{R}^{N}, i=1,\ldots,9$, denotes the discretized vector of $\boldsymbol{x}_i$;
$\boldsymbol{1}$ is an all-one vector;  $\boldsymbol{.}^2$ denotes element-wise square;
$\boldsymbol{D}\in\mathbb{R}^{N\times N+1}$ is the differential approximation matrix;
$\displaystyle\boldsymbol{\xi_{0,i}}$ denotes $i$-th component of $\boldsymbol{\xi_{0}}$; for simplicity, the multiplication of two vectors denotes the Hadamard product $\odot$, which is the element-wise product.

Thus, the transcribed problem is still a nonlinear programming problem and can be solved using any Non-linear Programming (NLP) solver. The calculated control input will then be sent to the quadrotor to drive the quadrotor to target the moving animal.
\subsubsection{Boundary Value Problem}
This method transforms the optimal problem into a boundary value problem via the calculus of variations. Then the boundary value problem can be solved (often numerically) for extremal trajectories. 

However, due to the severe nonlinearities, this method cannot be directly applied to solve the original optimal control problem in Eq.\eqref{function}. It should be further assumed that the system is in the near hovering state ($\theta=0, \phi = 0$), such that $d_z\approx(-\boldsymbol{x}_1\boldsymbol{u}_2-\boldsymbol{x}_3)$. In this case, the optimal problem can be reformulated as
\begin{equation}
\begin{aligned}
\min \quad & \displaystyle \mathcal{J}^{'}=\int_{0}^{{{t}_{f}}} G^{'}[\boldsymbol{x}(t), \boldsymbol{u}(t),t]\text{d}t\\
\textrm{s.t.} \quad   &     \displaystyle\boldsymbol{\dot{x}}(t)=\boldsymbol{h}^{'}[\boldsymbol{x}(t),\boldsymbol{u}(t)],\boldsymbol{x}(0)=\boldsymbol{{\xi}_{0}}  \\
\end{aligned}
\end{equation}
where $\displaystyle G^{'}={\frac{1}{2}\boldsymbol{u}^T\boldsymbol{u}+k_{1}{{\left( d_x+3 \right)}^{2}}+k_{2}d_y^2+k_{3}d_z^2}$; $ \boldsymbol{h}^{'}=[{\boldsymbol{x}_{4-6}}^T,g{\boldsymbol{u}_2}-c_1\boldsymbol{x}_7,-g\boldsymbol{u}_3-c_2\boldsymbol{x}_8,\boldsymbol{u}_1-c_3\boldsymbol{x}_9-g,g{\boldsymbol{u}_2}-c_1\boldsymbol{x}_7,-g\boldsymbol{u}_3-c_2\boldsymbol{x}_8,\boldsymbol{u}_1-c_3\boldsymbol{x}_9-g]^T$.
Firstly, by applying the Lagrange multiplier method, the augmented performance objective can be constructed as 
\begin{equation}
\displaystyle \mathcal{J}'=\int_{0}^{{{t}_{f}}} G^{'}[\boldsymbol{x}(t), \boldsymbol{u}(t),t]+\boldsymbol{\lambda}^T\left[\boldsymbol{h}^{'}(\boldsymbol{x}(t),\boldsymbol{u}(t))-\dot{\boldsymbol{x}}(t)\right]\text{d}t
\end{equation}
where $\boldsymbol{\lambda}\in\mathbb{R}^9$ is the costate vector. Then, via the calculus of variations, the optimal control input can be generated as 
\begin{equation}
\begin{aligned}
\boldsymbol{u}^*(t) &= \boldsymbol{V}(\boldsymbol{x},\boldsymbol{\lambda})\\
&=\left[-\boldsymbol{\lambda}_6-\boldsymbol{\lambda}_9, \frac{-(\boldsymbol{\lambda}_4+\boldsymbol{\lambda}_7)g-2k_3\boldsymbol{x}_1\boldsymbol{x}_3}{2k_3\boldsymbol{x}_1^2+1},g(\boldsymbol{\lambda}_{5}+\boldsymbol{\lambda}_{8})\right]^T
\end{aligned}
\end{equation}
in which $\boldsymbol{\lambda}_i$ denotes $i$-th component of $\boldsymbol{\lambda}$; $\boldsymbol{\lambda}$ and $\boldsymbol{x}$ can be calculated by solving the following boundary value problem via well-developed numerical algorithms.
\begin{equation}\label{costate}
\begin{aligned}
&\dot{\boldsymbol{\lambda}}(t)=\boldsymbol{S}\left[\boldsymbol{x}(t),\boldsymbol{\lambda}(t)\right], \boldsymbol{\lambda}(t_f)=\boldsymbol{0}_{9\times1}\\
&\dot{\boldsymbol{x}}(t)=\boldsymbol{h}^{'}\left[\boldsymbol{x}(t)\boldsymbol{\lambda}(t)\right], \boldsymbol{x}(0)=\boldsymbol{{\xi}_{0}}
\end{aligned}
\end{equation}
where $\boldsymbol{S}=\left[-2k_{1}\left(\boldsymbol{x}_{1}+3\right)+2k_{3}d_z \boldsymbol{u}^*_{2},-2k_2\boldsymbol{x}_{2},2 k_{3}d_z, -\boldsymbol{\lambda}_{1-3}^T\right.,$\\
$\left.(\boldsymbol{\lambda}_{4-6}^T+\boldsymbol{\lambda}_{7-9}^T)\boldsymbol{C}\right]^T$, $\boldsymbol{u}^*_i$ is $i$-th component of $\boldsymbol{u}^*$.
\subsection{Evaluation Environment Setup}
The simulation is implemented on a desktop computer with an Intel Core i7-10700 CPU running at 2.9GHz, with 16GB of RAM. It runs on a Linux (Ubuntu 18.04) operating system and is constructed in a physically realistic environment. 

\begin{figure}
	\centering
	\includegraphics[width=1.0\linewidth]{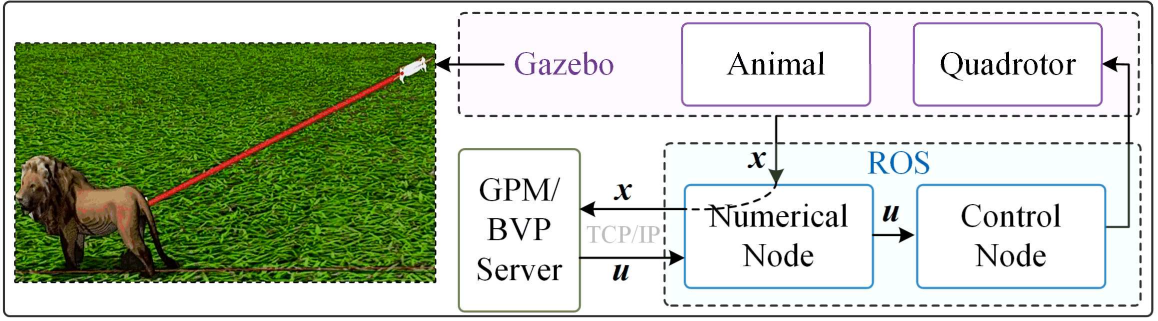}
	\caption{The schematic diagram of system structure in simulation}
	\label{fig:system structure}
\end{figure}

The implemented structure of this simulation environment is illustrated in Fig. \ref{fig:system structure}. The quadrotor and animal dynamics are implemented in the Gazebo environment. The numerical node is utilized to calculate the optimal control input via the received state in Python. The control node is implemented in C++ to calculate the optimal control input and then control the quadrotor at a frequency of 50 Hz. 
Both control node and numerical node are implemented within the Robot Operating System (ROS). 

To further increase the computational efficiency of the generic optimizers, the following three acceleration techniques are adopted. 

1) The optimal control computations of generic optimizers are separately running on GPM/BVP server, which is accelerated by Numba 0.51.2 of Python.
In such a case, the numerical node of generic optimizers only serves to forward the message and is bridged with GPM/BVP server by TCP/IP.

2) The calculations of the generic optimizers are combined with warm-starting \cite{Wang2010fast}, i.e., initializing using the estimate from the previous calculation results.

3) In the following simulations and experiments, to reduce the computing complexity of the generic optimizers, the position of the quadrotor along $\boldsymbol{e}_2^\mathcal{I}$ is separately controlled by PD controller as the proposed EER does. 

In all of the following simulation tests, the EER and generic optimizers are implemented as follows:

1) EER: with the trial and error approach, the two weighting matrices in the EER are selected as $\boldsymbol{Q}_1=\text{diag}\{58,264,30,10\},  \boldsymbol{Q}_2=\text{diag}\{40,30\}$. The position control gains along $\boldsymbol{e}_2^\mathcal{I}$ are $k_p = 2.0, k_d = 3.0$. 

2) Generic optimizers: the terminal time is selected as $t_{f} = 2$; the weighting gains are chosen as $ k_{1} = k_{3} = 50$; the number of LG points in GPM is chosen as $ N = 7$. The position control gains along $\boldsymbol{e}_2^\mathcal{I}$ are the same as the EER. Boundary value problem in BVP and the NLP in GPM are solved with scipy.integrate.solve$\_$bvp and scipy.minimize in Python 3.6, respectively. 

\subsection{Simulation Tests}
In this section, two cases are carried out on the desktop computer to investigate the targeting performance and computational efficiency of the designed method. 

\begin{figure}
	\centering
	\subfigure[Targeting error]{\includegraphics[width=0.49\linewidth]{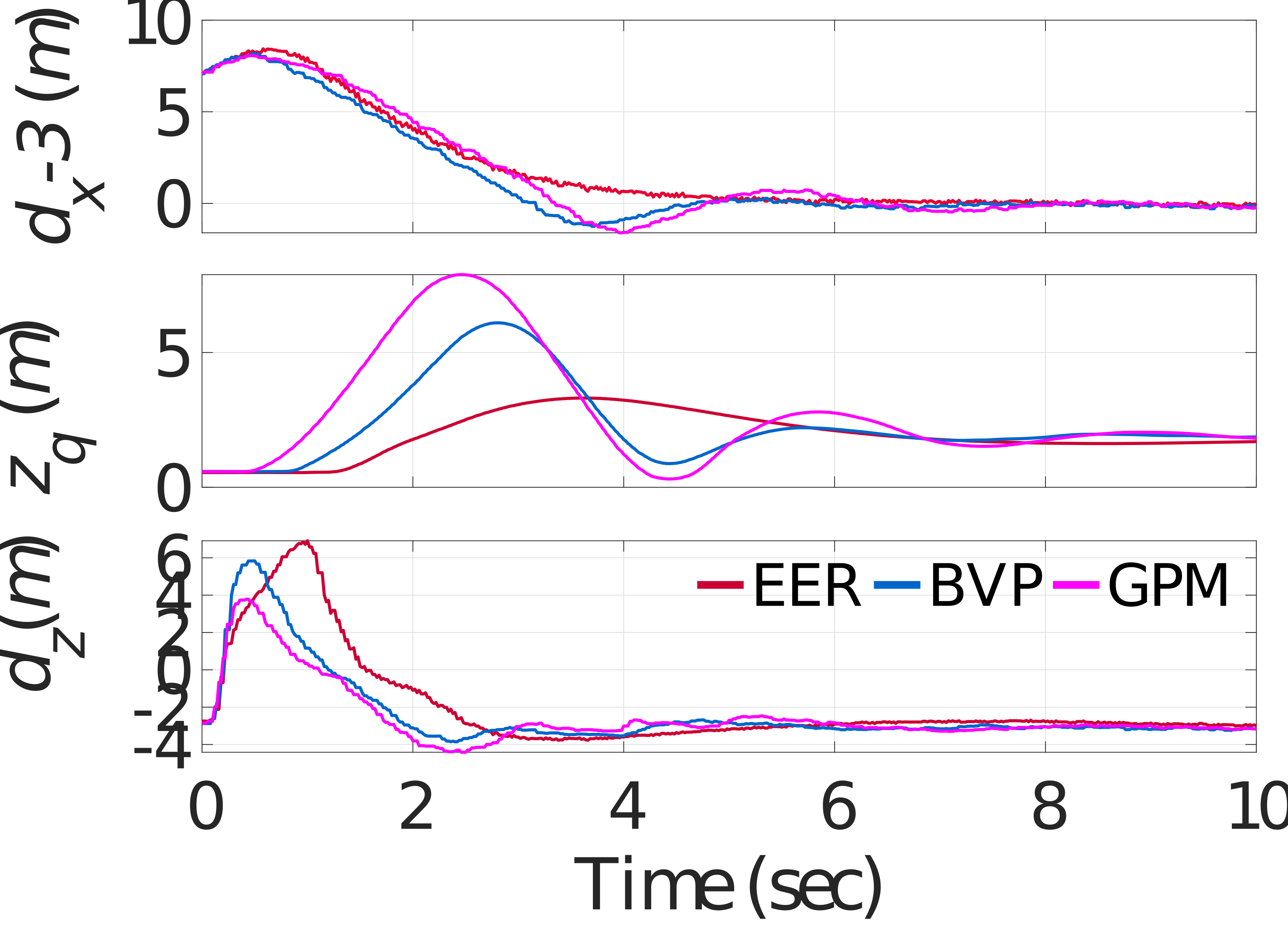}}
	\subfigure[Control inputs and cost time]{\includegraphics[width=0.49\linewidth]{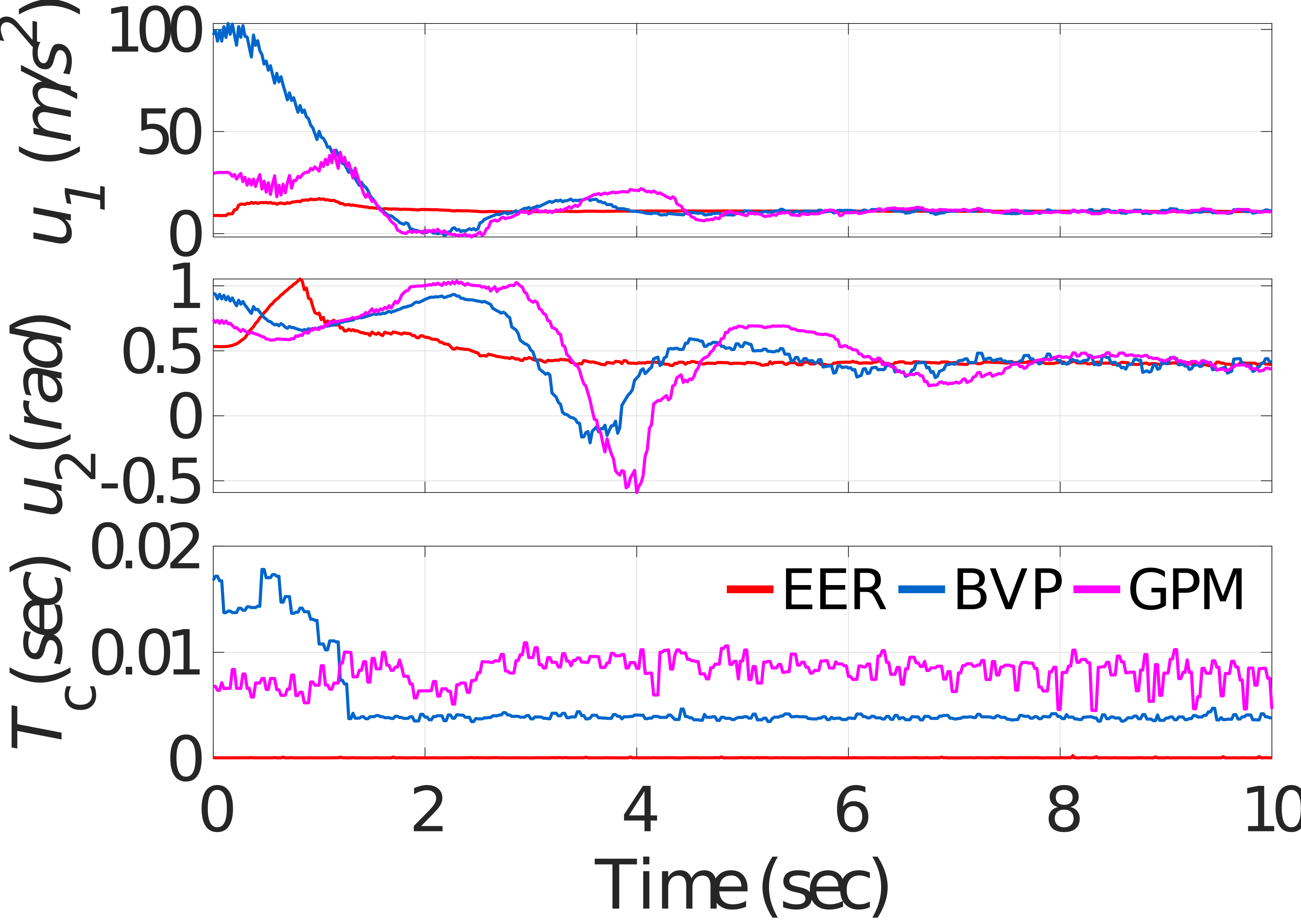}}
	\caption{Case 1 in simulation: the animal moves by constant speed}
	\label{fig:Case1}
\end{figure}

Case 1:
A first test is conducted, where the animal is moving along $\boldsymbol{e}_1^{\mathcal{I}}$ at a constant speed, i.e., $v_{tx} = 3\,\text{m/s}$, and the height of the targeted-point is 0.61m.
At the initial moment, the animal is located at the origin and the quadrotor is hovering at $[-10, 0, 0.61]^T$. The comparative targeting results are depicted in Fig. \ref{fig:Case1}, where $T_c$ represents the computation time in each controller updating timestep. Because the positions along $\boldsymbol{e}_2^\mathcal{I}$ of the compared methods are all controlled by the PD controller, there exists little difference in targeting performance between different methods in $d_y$. Therefore, $d_y$ is omitted in the figure. The average computation times of EER, BVP and GPM are 0.07 ms, 3.88 ms, 8.40 ms, respectively. It can be seen that the computational efficiency of the EER is on the order of 120 times faster than GPM. The result can be interpreted as follows. For the generic optimizers, severe nonlinearities of the considered problem deteriorate the computational efficiency. In contrast, for the EER, the severe nonlinear characteristics is peeled off and only a quadratic optimal control problem is required to be solved. In such a case, the control input is generated analytically, which provides excellent profits for high computational efficiency.

Furthermore, it can also be noted that large overshooting and strong oscillation of $z_q$ appears in generic optimizers. While with the EER, the height of the quadrotor $z_q$ converges relatively smoothly and fast without overshooting. The reasons for this phenomenon lie in two aspects. Firstly, for the EER, thanks to its high computational efficiency, the control action can be updated in real-time and thus presents a smooth characteristic as shown in Fig. \ref{fig:Case1}. In contrast, for generic optimizers, the long computation time leads to the trembling and discontinuity of the control inputs, deteriorating the targeting performance. Secondly, for the generic optimizers, according to Eq. \eqref{function}, the vertical position of the quadrotor is coupled with the translational position as well as the pitch angle in the performance objective. Therefore, $z_q$ cannot be constrained directly by the performance objective. On the contrary, the pitch angle and the translational position decoupled in the EER. Therefore, the vertical position of the quadrotor can be well constrained by the performance objective in a reasonable range. In this case, no oscillation, smaller overshooting as well as fast convergence appear in $d_x, d_z, z_q$ when utilizing the proposed EER to target the animal.

% \begin{figure}
% 	\centering
% 	\includegraphics[width=0.8\linewidth]{fig/3_case1}
% 	\caption{Simulation result: case 1}
% 	\label{fig:Case1}
% \end{figure}
 \begin{figure}
	\centering
	\subfigure[Targeting error]{\includegraphics[width=0.49\linewidth]{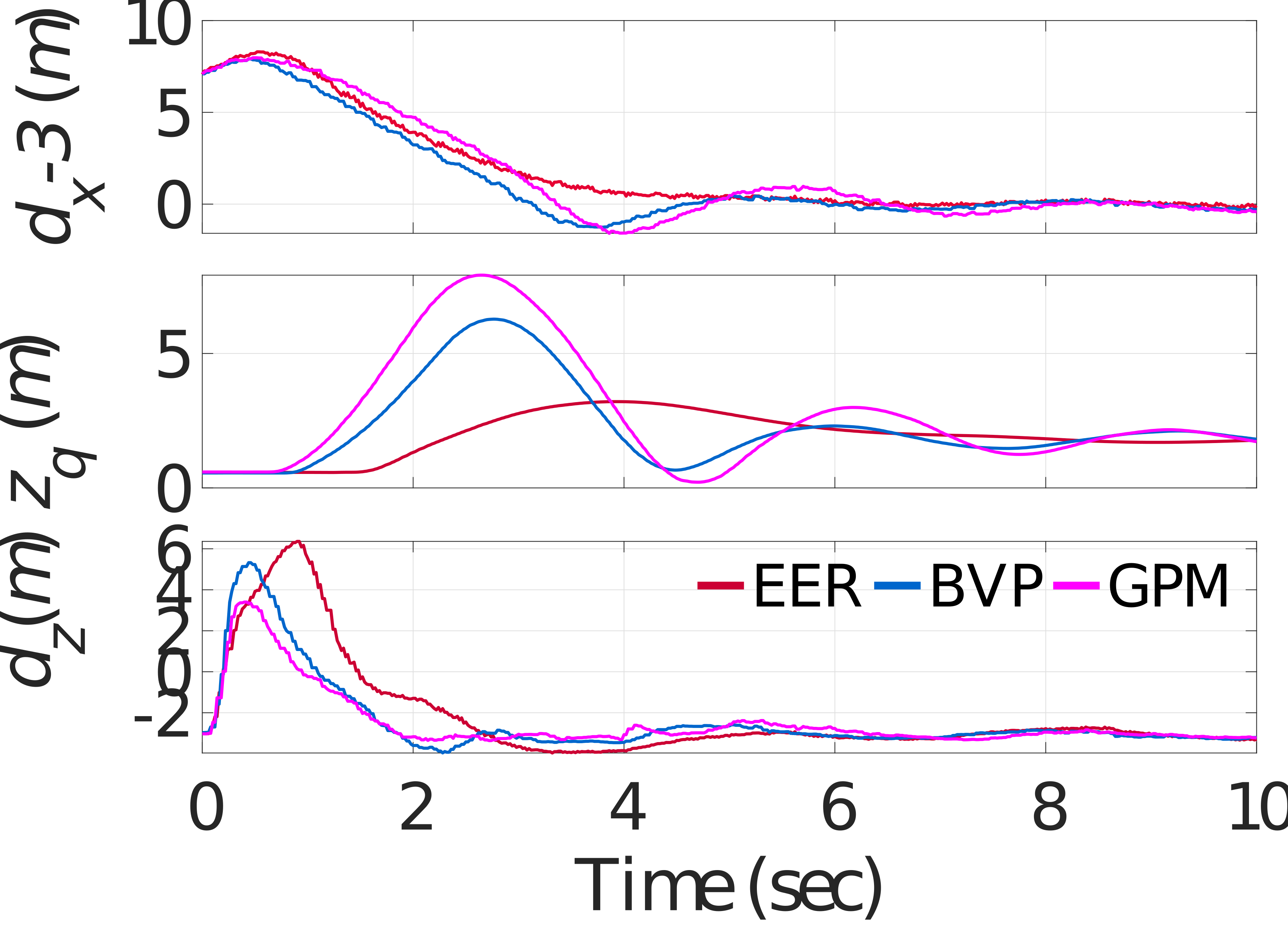}}
	\subfigure[Control inputs and cost time]{\includegraphics[width=0.49\linewidth]{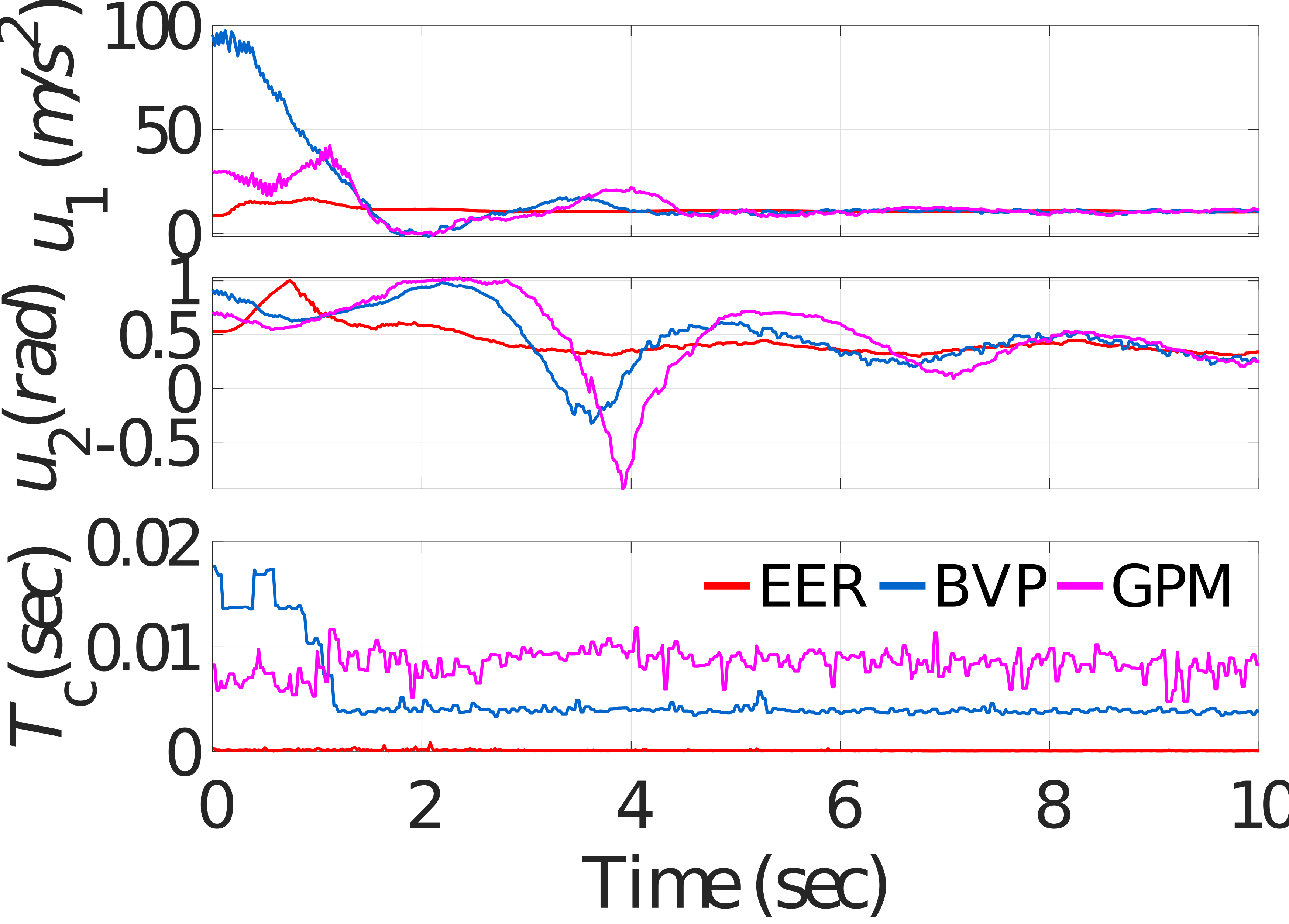}}
	\caption{Case 2 in simulation: the animal moves by time-varying speed}
	\label{fig:Case2}
\end{figure}
% \begin{figure}
% 	\centering
% 	\includegraphics[width=0.8\linewidth]{fig/3_case2}
% 	\caption{Simulation result: case 2}
% 	\label{fig:Case2}
% \end{figure}

Case 2:
The second test is then carried out when the velocity of the animal is time-varying. Specifically, the animal's speed is sinusoidal at a frequency of 0.5 Hz, ranging from 2.6 to 3. Similar to case 1, the quadrotor is hovering at $[-10, 0, 0.61]^T$ and the animal is located at the origin initially. The targeting result is illustrated in Fig. \ref{fig:Case2}. In this case, the average computation times of EER, BVP, and GPM are 0.1 ms, 3.95 ms, 8.57 ms, respectively. It can be seen that the computational efficiency of the EER is on the order of 85.7 times faster than GPM. It should be mentioned that the computation time of the same approach under the two cases is slightly different. This fluctuation in computation time is affected by some other parallel tasks executing on the desktop computer.

Additionally, in steady-state, the oscillations of $z_q$ in the generic optimizers are larger than that in case 1. By comparison, for the EER, the vertical position of the quadrotor could also provide comparatively excellent performance. This result can be interpreted as follows. Thanks to the high computational efficiency of the proposed EER, it is reasonable to treat the animal's velocity as a constant at each controller updating timestep in the system modeling. Therefore, the time-varying characteristic of the animal's velocity has little effect on targeting performance in the EER. For the generic optimizers, due to the long computation time, assuming the animal velocity as a constant will lead to large modeling errors and worsen the targeting performance when the animal speed is time-varying.

\section{Experiment}
In order to testify the feasibility of EER in practical application, real-time mimic biological experiments are carried out in this section.

\subsection{Experimental Setup}
\begin{figure}
	\centering
	\includegraphics[width=0.8\linewidth]{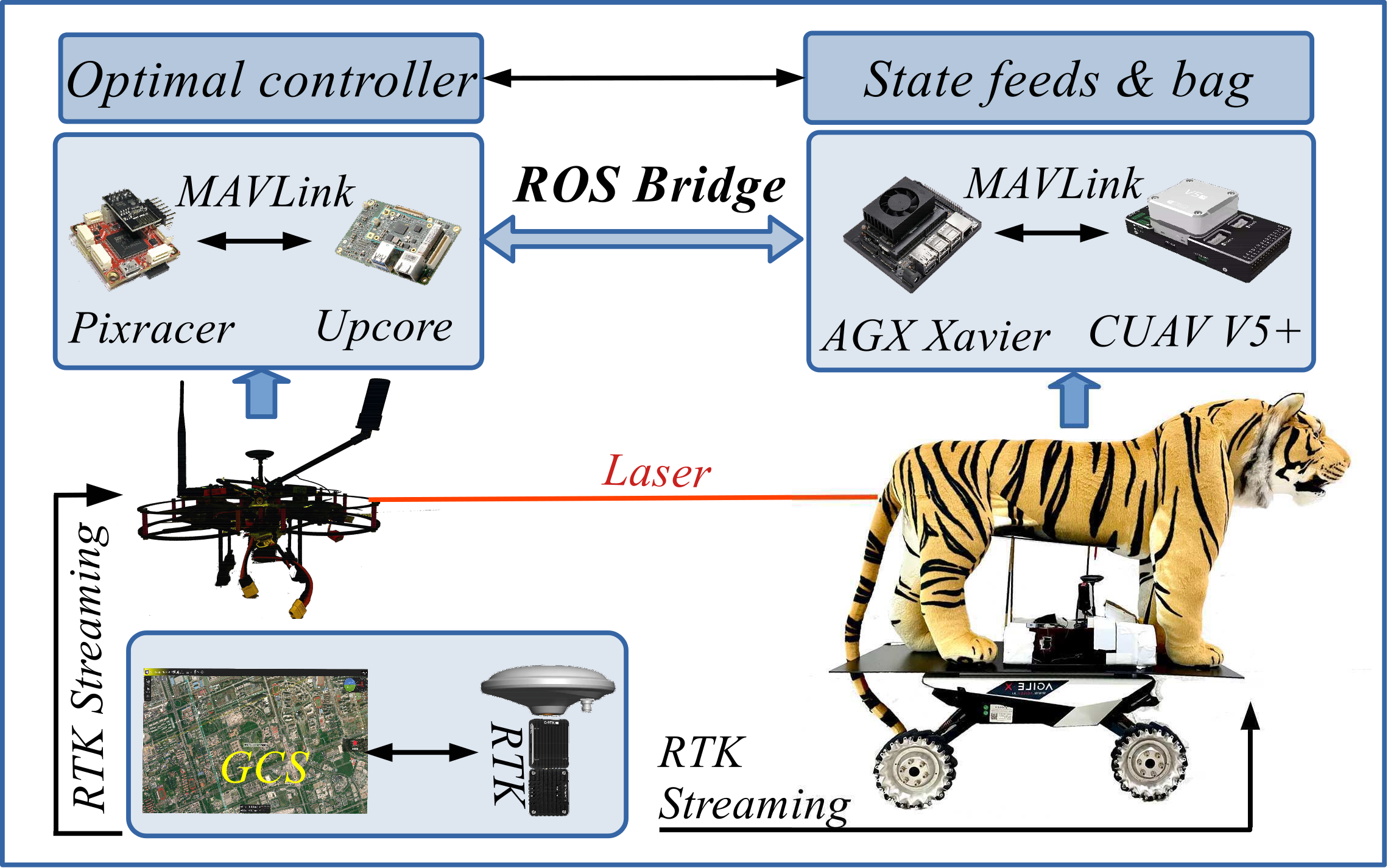}
	\caption{Mimic biological experimental testbed}
	\label{fig:hardware_structure}
\end{figure}
A mimic biological experimental test-bed is developed to verify the developed EER for the targeting task, as shown in Fig. \ref{fig:hardware_structure}. This test-bed consists of a quadrotor, a targeted animal and a Ground Control System (GCS). The airframe is a custom-developed quadrotor that weighs 1.98kg. Its attitude is controlled with an open-source controller Pixracer$^\circledR$, which mainly adopts the cascaded PID control technique. All control methods are implemented in an onboard computer, namely the Up-core$^\circledR$, and it communicates with the Pixracer$^\circledR$ by MAVLink. 
The targeted animal is mounted on a small car, which is controlled by a remote controller. 

The position and velocity of the quadrotor, as well as the targeted animal, is estimated by the Real-time Kinematic positioning(RTK). The RTK is a differential GNSS technique which provides centimeter accuracy in the vicinity of a base station. In addition, an LIDAR-Lite v3$^\circledR$ is implemented to estimate the altitude of the quadrotor accurately. The quadrotor obtains the position of the targeted animal via a ROS bridge. The detailed implementations of the positioning system of the targeted animal are as follows: 1) The Ground Control System (GCS) first receives the messages from RTK, and then transmits them to CUAV V5+$^\circledR$ by radio. 2) Then, the CUAV V5+$^\circledR$ accepts the messages, and sends them to the NVIDIA$^\circledR$ Jetson AGX Xavier by MAVROS. 3) AGX Xavier publishes the position and velocity of the targeted animal by ROS, and the ROS implemented on the Up-core of the quadrotor can subscribe to the messages.

In all of the following experiments, with a trial and error approach, the two weighting matrices regarding the developed EER are tuned as follows: $\boldsymbol{Q}_1=\text{diag}\{116,441,87,18\},  \boldsymbol{Q}_2=\text{diag}\{40,30\}$. The control gains of BVP and GPM are selected as $ k_{1} = k_{3} = 2.5$. The selection of the other parameters in optimizations is consistent with the simulation.
\subsection{Experimental Results}
\begin{figure}
	\centering
	\includegraphics[width=0.95\linewidth]{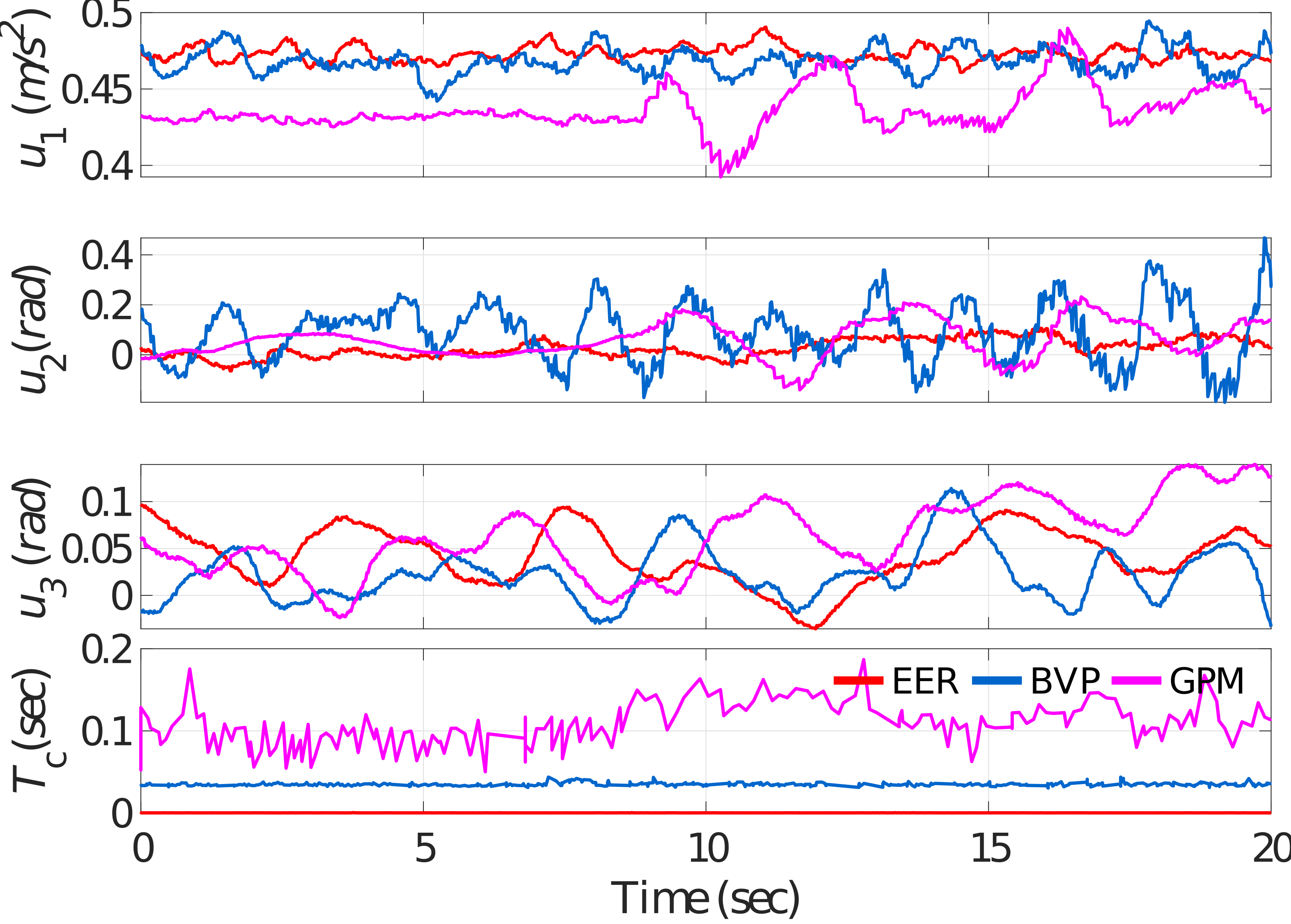}
	\caption{Case 1 in experiment: control inputs and computation time }
	\label{fig:controlinput}
\end{figure}
\begin{figure}
	\centering
	\subfigure[Targeting error]{\includegraphics[width=0.47\linewidth]{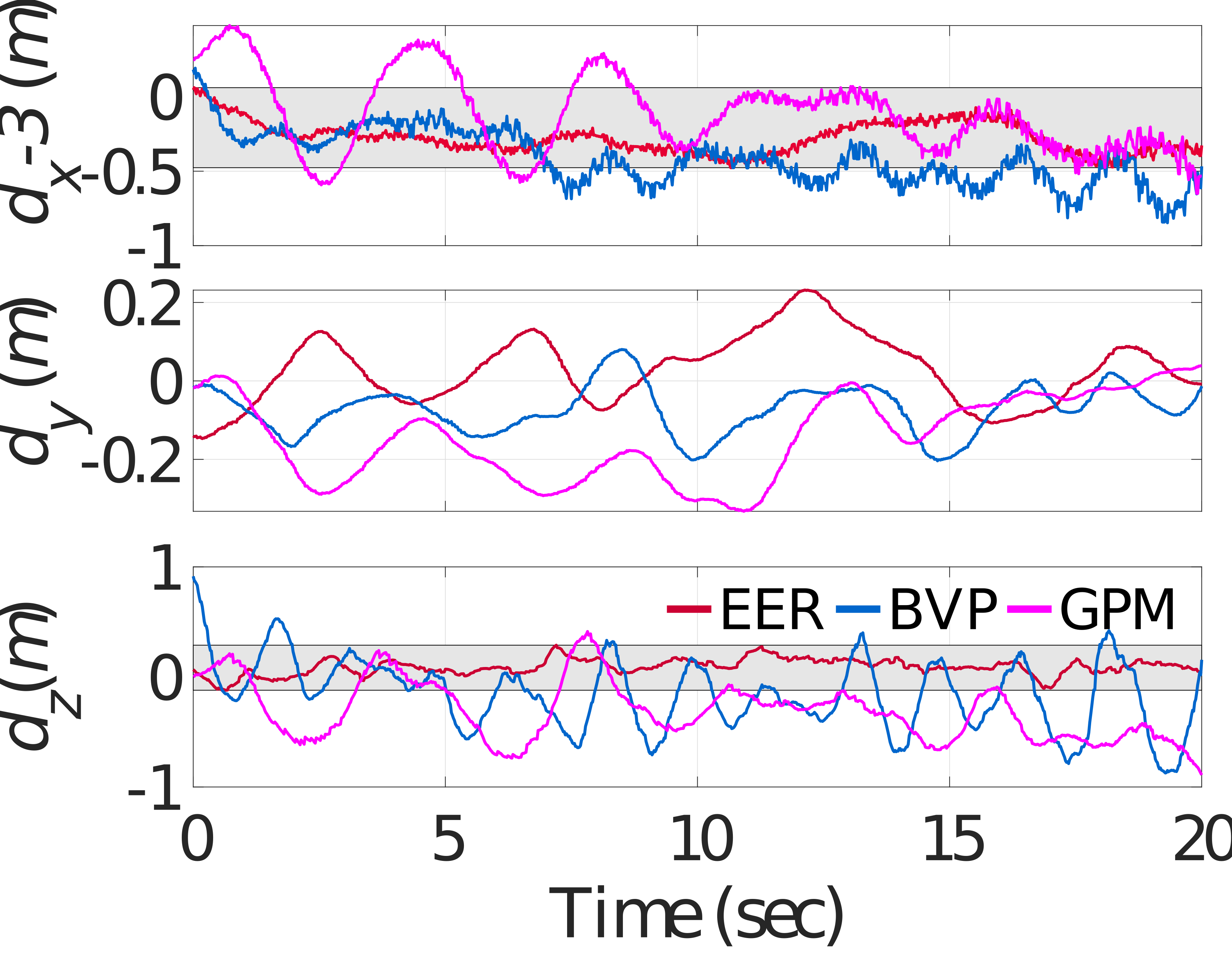}}
	\subfigure[Mean absolute error]{\includegraphics[width=0.49\linewidth]{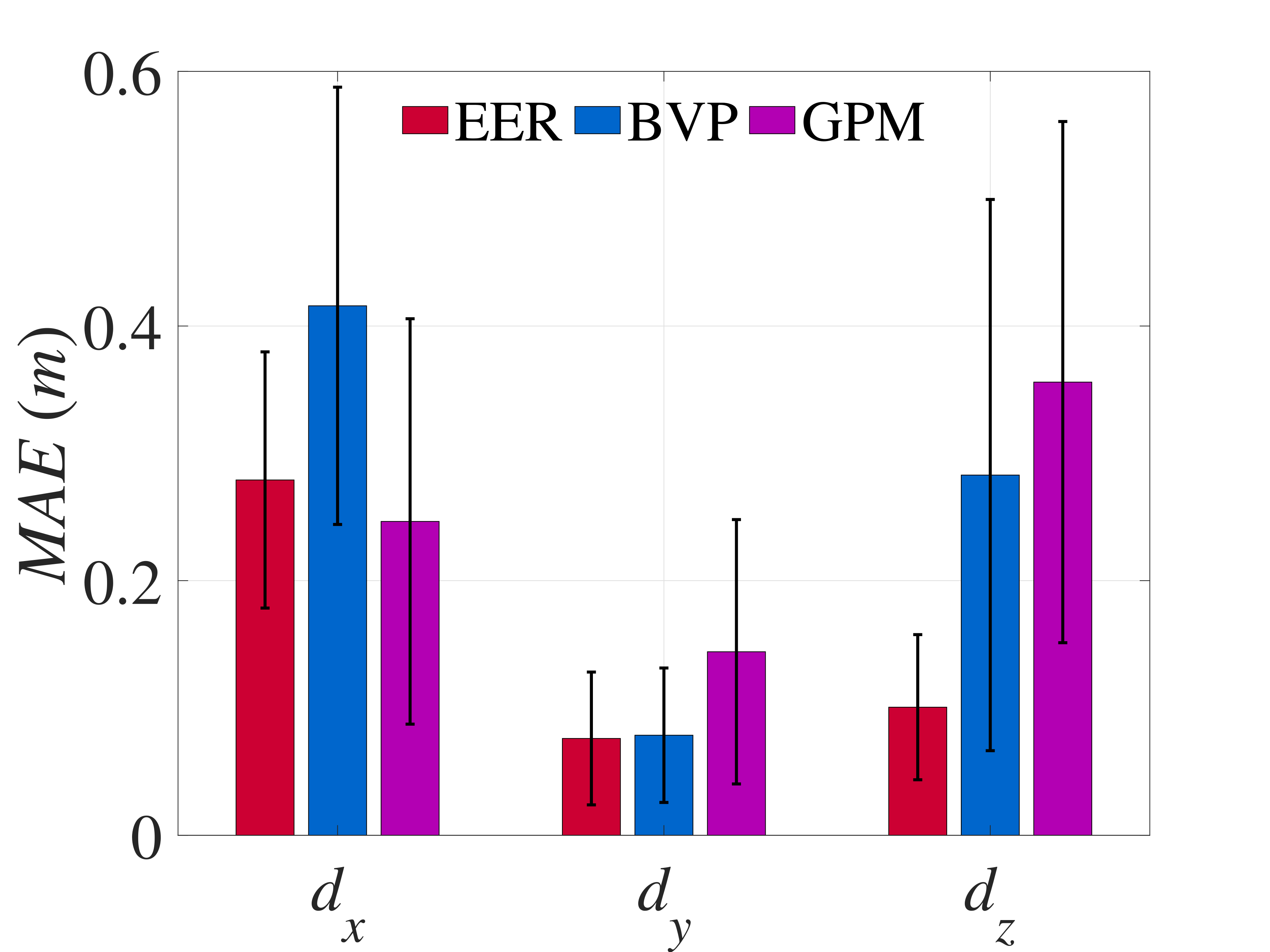}}
	\caption{Case 1 in experiment: targeting error}
	\label{fig:4_targeterror}
\end{figure}

A first test is conducted, where the animal is located at the origin and the quadrotor is hovering at $[-3, 0, 0.7]^T$ at the initial moment. The height of the desired point is 0.7 m and it then moves along $\boldsymbol{e}_1^{\mathcal{I}}$. Its speed increases from 0 $m/s$ to about 1.5 $m/s$ with nearly a uniform acceleration. The comparative targeting results are depicted in Fig. \ref{fig:controlinput} and Fig. \ref{fig:4_targeterror}. 

Fig. \ref{fig:controlinput} depicts the control inputs and the computational time of the EER, BVP, and GPM.
In this case, the average computation times of EER, BVP, and GPM are 0.27 ms, 38.12 ms, 108.87 ms, respectively. It can be seen that the EER can compute the control action on the order of 400 times faster than GPM on Up Core$^\circledR$ at least. Particularly, EER can achieve a control frequency of thousands of hertz. 
Compared with the simulation, we can find that the computational efficiency of the EER is stabler than the generic optimizers on different platforms. To be specific, the computation time of the EER in the experiment is similar to that in simulation, while the computation time of the BVP and GPM is about tens or even hundreds of milliseconds slower than in simulation.

Fig. \ref{fig:4_targeterror} illustrates the targeting error and the mean absolute error (MAE) of the three compared methods. 
Compared with the simulation, it can be inferred that in the experiment, the superiority of EER compared to the generic optimizers regarding targeting performance is more obvious.
% Specifically, in the outdoor experiment, the targeting performance of the EER is far better than that of generic optimizers. 
Specifically, the MAE of $d_z$ with EER are smaller than those with the generic optimizers. In addition, regarding $d_x, d_z$, EER presents relatively smaller oscillation than generic optimizers, as shown in Fig. \ref{fig:4_targeterror}(a). This phenomenon can be interpreted as follows. 
Compared to the Gazebo environment, there exists inherent time delay and various unknown disturbances in the real system, and the processing capability of the onboard computer is not as good as that of the desktop computer in the simulation. Besides, the fluctuation of the animal's velocity is larger than that of the simulation, leading to a larger modeling error in generic optimizers. In such a case, the computational efficiency in the real experiment is a more crucial issue than that in the simulation.
However, due to severe nonlinearities of this optimal problem, generic optimizers cannot solve this problem with sufficient computational efficiency on the onboard computers, leading to a long time delay in the state feedback. Therefore, 
the targeting performance of the generic optimizers is worse than that of the proposed EER. The above results effectively demonstrate the significance of the computational efficiency for this targeting problem.

% the time delay and discontinuity of the control input in the control process, leading to the oscillation of the generic optimizers. 
% Secondly, when the computing efficiency is low, the assumption that the animal's velocity is a constant brings a large modeling error, further worsening the targeting performance.
\begin{figure}
	\centering
	\includegraphics[width=0.9\linewidth]{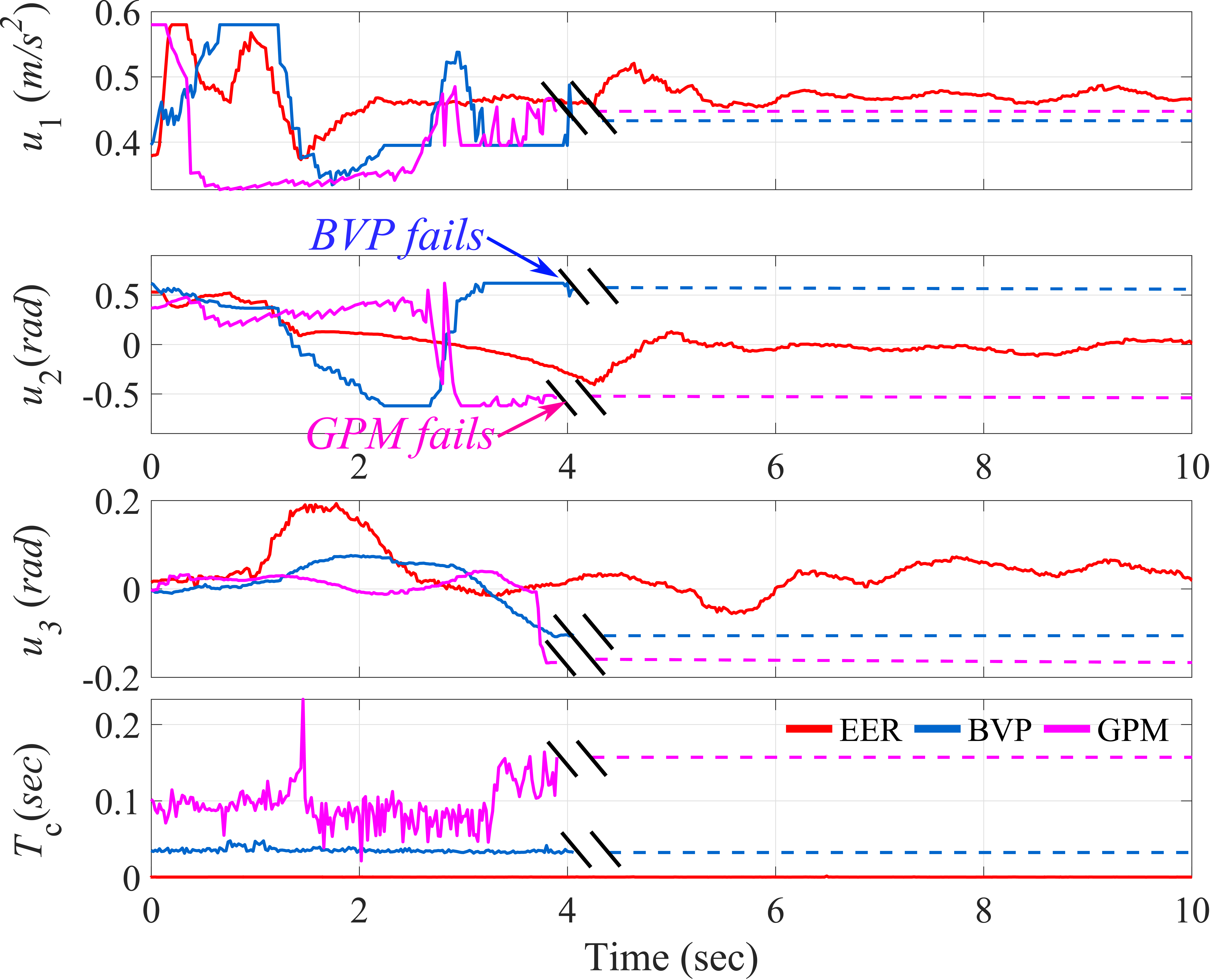}
	\caption{Case 2 in experiment: control inputs and computation time }
	\label{fig:4_case2_controlinput}
\end{figure}

\begin{figure}
	\centering
	\includegraphics[width=1.0\linewidth]{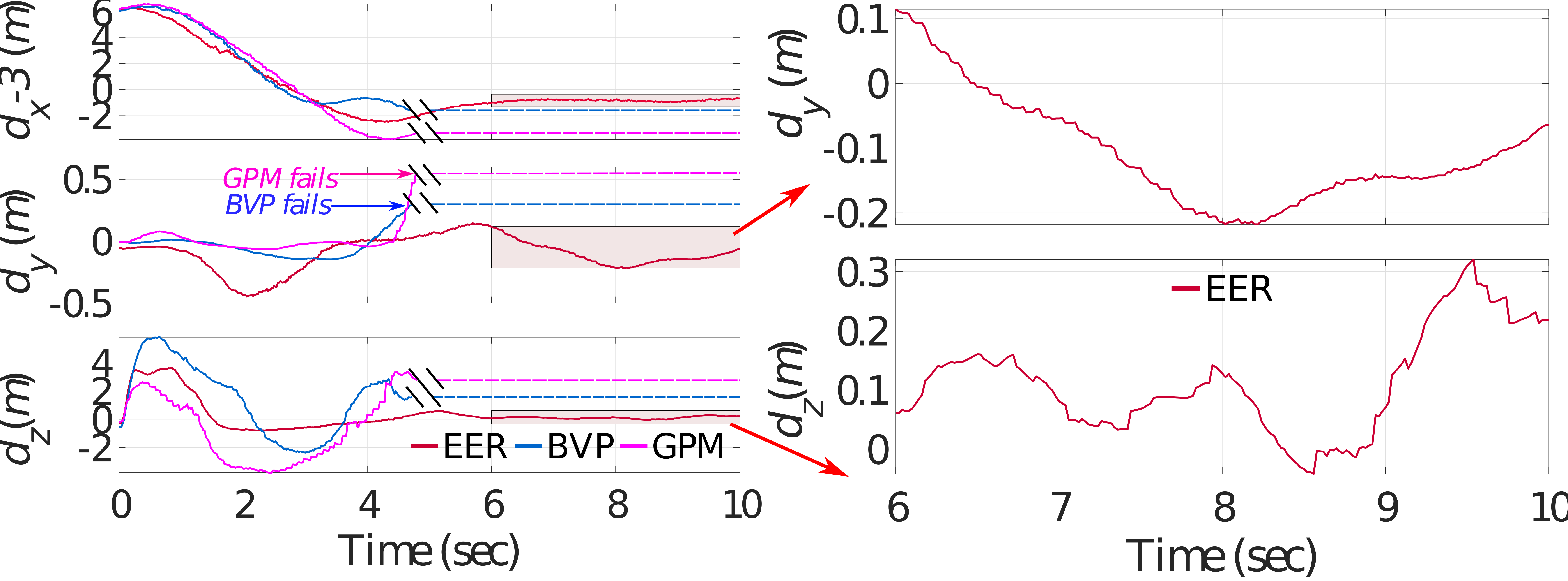}
	\caption{Case 2 in experiment: targeting error}
	\label{fig:4_case2_error}
\end{figure}

The second test is conducted, where the quadrotor is targeting a moving animal at a long distance. The animal is moving at speed close to 1.0 $m/s$ along $\boldsymbol{e}_1^{\mathcal{I}}$ and is 9 $m$ away from the quadrotor. In the experiment, it can be observed that with the generic optimizers, the quadrotor flies very high and oscillates up and down, finally directly collides the animal or crash into the ground. 
Fig. \ref{fig:4_case2_controlinput} and Fig. \ref{fig:4_case2_error} depict the comparative results, where the double slash denotes the crash time.

In Fig. \ref{fig:4_case2_controlinput}, the control input and computation time of the proposed EER and generic optimizers are illustrated. The average computation time of EER, BVP, and GPM from 0 to 4.60 $s$ are 0.29 ms, 35.40 ms, 100.1 ms, respectively. It can be seen that the computational efficiency of the EER is on the order of 345 times faster than GPM. It can be seen that the generic optimizers fail around 4.6 s. This is because the tiger has speed at the initial moment and the initial error is large. Therefore, the targeting task in this case has higher real-time requirements. Due to the low computational efficiency of the generic optimizers, they cannot adjust the targeting error in time and leads to the crash.

Fig. \ref{fig:4_case2_error} shows the compared targeting error. It should be stated that the targeting performance in $d_y$ and $d_z$ is mainly concerned, so the zoomed views of them are given particularly. Even in such a case, with the EER, the quadrotor quickly catches up with the animal with small oscillation and overshooting. Furthermore, after the system reaches a steady state, the proposed presents excellent targeting performance again. Particularly, the targeting error of $d_y$ is between -0.2 and 0.1, and the targeting error of $d_z$ is between -0.1 and 0.3, which can ensure the completion of the targeting task.

All the aforementioned simulation and experimental results demonstrate that the proposed EER has high computational efficiency either on a powerful desktop computer or an onboard computer and can achieve the best targeting performance than those of the generic optimizers.
% Additionally, EER can achieve a control frequency of thousands of hertz on a commonly utilized onboard computer and thus ensure computational efficiency in real-time.
\section{Conclusion}
In this paper, in order to continuously target a moving animal with a highly underactuated quadrotor, an efficient egocentric regulator is proposed. This strategy directly constructs the optimal tracking problem in an egocentric manner with regard to the quadrotor's body coordinates and peels off the system nonlinearities through a mapping of the feedback states as well as control inputs, between the inertial and body coordinates. In this way, only a quadratic performance objective with linear constraint is required to be solved with an analytic evaluation. To verify the advantages and computational efficiency of the proposed strategy, mimic biological simulations and experiments are first carried out. Results demonstrate that the proposed EER can achieve faster convergence and smaller overshooting than those of the generic optimizers and its computational efficiency is highest and stablest than generic optimizers on different platforms. Particularly, on the onboard computer, the computation time of the EER is about 0.3 ms, which is on the order of 350 times faster than generic nonlinear optimizers and can accomplish a control frequency of thousands of hertz. In view of these convincing results, the development of this work is able to effectively enhance the computational efficiency for continuous targeting problems of quadrotors in practice.

%\section*{Acknowledgement}
%This work is partially supported by the Scientific and Technical Innovation 2030 - “Artificial Intelligence of New Generation” Major Project (2018AAA0102704) and National Natural Science Foundation of China (Grant No. 51975348, 51605282).

\ifCLASSOPTIONcaptionsoff
  \newpage
\fi

\normalem 
\bibliographystyle{IEEEtran}
\bibliography{ref}

\vspace{-1cm}

\begin{IEEEbiography}
[{\includegraphics[width=1in,clip]{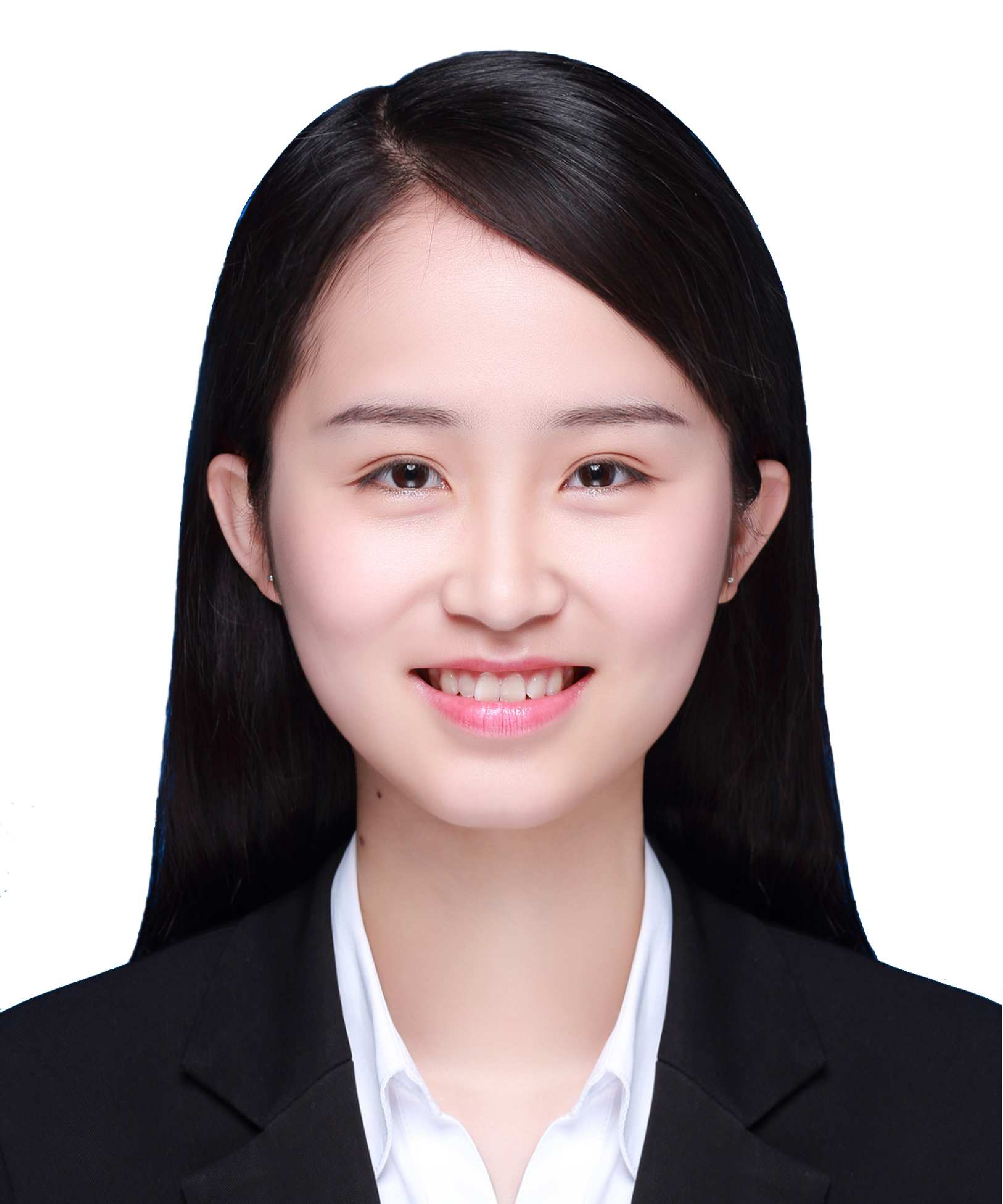}}]{Ziying Lin}
received the B.S. degree in mechanical engineering from Nanjing University of Science and Technology, Nanjing, China, in 2019. She is currently a Ph.D. candidate in the School of mechanical and engineering, Shanghai Jiao Tong University, China. Her research focuses on cooperation and control of unmanned system.
\end{IEEEbiography}
\vspace{-1cm}
\begin{IEEEbiography}
[{\includegraphics[width=1in,height=1.2in,clip]{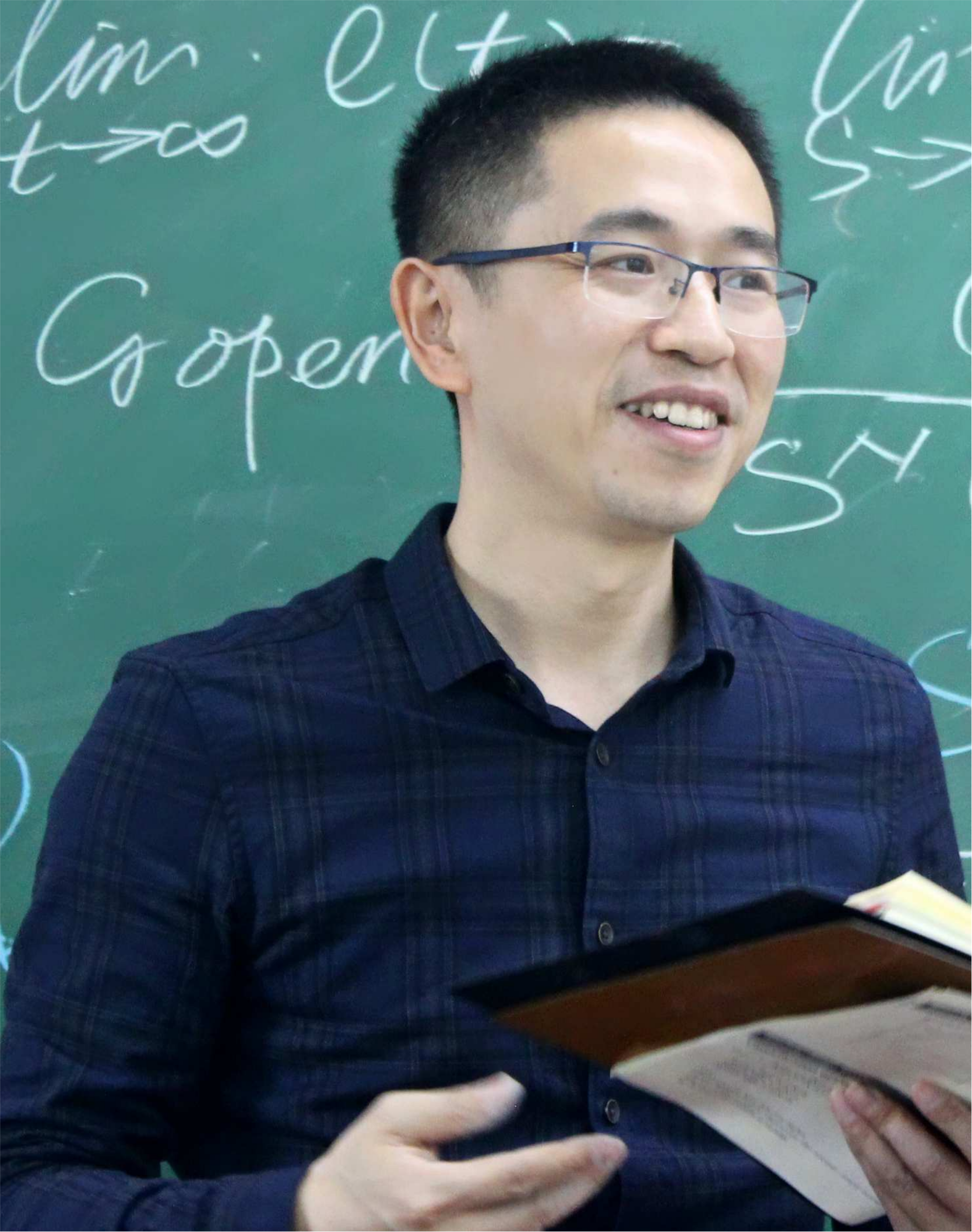}}]{Wei Dong}
received the B.S. degree and Ph.D. degree in mechanical engineering from Shanghai Jiao Tong University, Shanghai, China, in 2009 and 2015, respectively. He is currently an associate professor in the School of Mechanical Engineering, Shanghai Jiao Tong University. His research interests include cooperation, perception and agile control of unmanned system.
\end{IEEEbiography}
\vspace{-1cm}
\begin{IEEEbiography}
[{\includegraphics[width=1in,height=1.2in,clip]{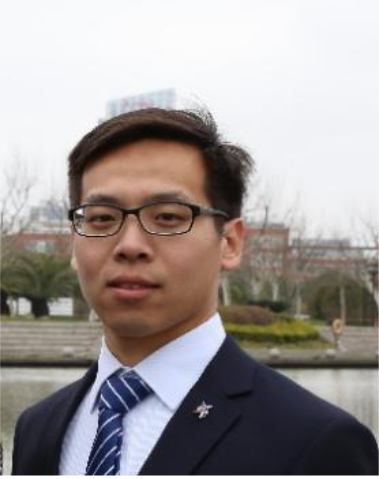}}]{Sensen Liu}
received the B.S. degree in mechanical engineering from Tongji University, Shanghai, China, in 2016. He is currently a Ph.D. candidate in the School of mechanical and engineering, Shanghai Jiao Tong University, China.His research focuses on aerial manipulation, gripper design, planning and control of unmanned aerial vehicles manipulation system.
\end{IEEEbiography}
\vspace{-1cm}
\begin{IEEEbiography}[{\includegraphics[width=1in,height=1.2in,clip]{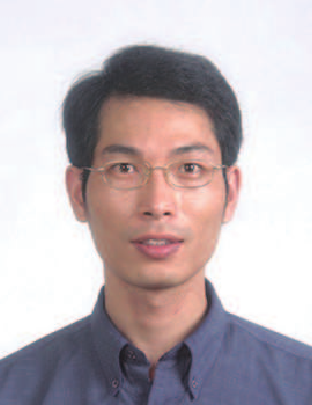}}]{Xinjun Sheng} received the B.Sc., M.Sc.,and Ph.D. degrees in mechanical engineering from Shanghai Jiao Tong University, Shanghai, China, in 2000, 2003, and 2014, respectively.
He is currently a Professor in the School of Mechanical Engineering, Shanghai Jiao Tong University. His current research interests include robotics, and bio-mechatronics. Dr. Sheng is a Member of the IEEERAS, the IEEEEMBS,and the IEEEIES.
\end{IEEEbiography}
\vspace{-1cm}
\begin{IEEEbiography}[{\includegraphics[width=1in,height=1.2in,clip]{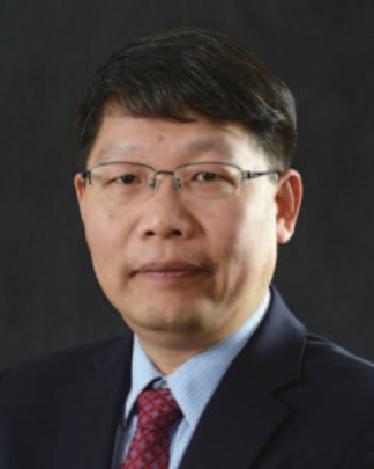}}]{Xiangyang Zhu}
received the B.S. degree from the Department of Automatic Control Engineering, Nanjing Institute of Technology, Nanjing, China, in 1985, the M.Phil. degree in instrumentation engineering and the Ph.D. degree in control engineering, both from Southeast Engineering, in 1989 and 1992, respectively. Since June 2002, he has been with the School of Mechanical Engineering, Shanghai Jiao Tong University, Shanghai, China, where he is currently a chair professor and the director of the Robotics Institute. His research interests include robotic manipulation planning, neuro-interfacing and neuro-prosthetics, and soft robotics. 
Dr. Zhu has received a number of awards including the National Science Fund for Distinguished Young Scholars from NSFC in 2005, and the Cheung Kong Distinguished Professorship from the Ministry of Education in 2007. He is serving on the editorial board of IEEE Transactions on Cybernetics and Journal of Bionic Engineering.
\end{IEEEbiography}

\end{document}